\pdfoutput=1

\documentclass[11pt]{article}

\usepackage[]{emnlp2023}

\usepackage{times}
\usepackage{latexsym}
\usepackage[normalem]{ulem}
\usepackage{booktabs}
\usepackage{amsmath}
\usepackage{amsfonts}
\usepackage{amsthm}
\usepackage{multirow}
\usepackage{multicol}
\usepackage{graphicx}
\usepackage{caption}
\usepackage[noabbrev]{cleveref}
\usepackage{subcaption}
\usepackage[T1]{fontenc}

\usepackage[utf8]{inputenc}

\usepackage{microtype}

\newcommand{\nospacetext}[1]{\makebox[0pt][l]{#1}}

\newcommand{\bert}[0]{\textsc{bert}}
\newcommand{\subj}[0]{\textsc{subj}}
\newcommand{\trec}[0]{\textsc{trec}}
\newcommand{\agn}[0]{\textsc{agn}}
\newcommand{\sst}[0]{\textsc{sst}}

\newcommand{\rnd}[0]{\textsc{rnd}}
\newcommand{\ent}[0]{\textsc{ent}}
\newcommand{\cs}[0]{\textsc{cs}}
\newcommand{\dal}[0]{\textsc{dal}}
\newcommand{\mc}[0]{\textsc{mc}}

\newcommand{\adapter}[0]{Adapter}
\newcommand{\lora}[0]{LoRA}
\newcommand{\pt}[0]{Prefix-tuning}
\newcommand{\uni}[0]{UniPELT}

\newcommand{\auc}[0]{AUC}

\title{Parameter-Efficient Language Model Tuning with\\ Active Learning in Low-Resource Settings}

\author{Josip Juki{\'{c}} \quad Jan {\v{S}}najder\\
TakeLab\\
Faculty of Electrical Engineering and Computing,  University of Zagreb, Croatia\\
\tt \{josip.jukic, jan.snajder\}@fer.hr
} 

\begin{document}
\maketitle


\begin{abstract}

Pre-trained language models (PLMs) have ignited a surge in demand for effective fine-tuning techniques, particularly in \textit{low-resource} domains and languages. Active learning (AL), a set of algorithms designed to decrease labeling costs by minimizing label complexity, has shown promise in confronting the labeling bottleneck. In parallel, adapter modules designed for parameter-efficient fine-tuning (PEFT) have demonstrated notable potential in low-resource settings. However, the interplay between AL and adapter-based PEFT remains unexplored. We present an empirical study of PEFT behavior with AL in low-resource settings for text classification tasks. Our findings affirm the superiority of PEFT over full-fine tuning (FFT) in low-resource settings and demonstrate that this advantage persists in AL setups.
We further examine the properties of PEFT and FFT through the lens of forgetting dynamics and instance-level representations, where we find that PEFT yields more stable representations of early and middle layers compared to FFT. Our research underscores the synergistic potential of AL and PEFT in low-resource settings, paving the way for advancements in efficient and effective fine-tuning.\footnote{Our code is available at \url{https://github.com/josipjukic/adapter-al}}

\end{abstract}

\section{Introduction}
\label{sec:intro}


Pre-trained language models (PLMs) have quickly become a staple in the field of natural language processing. With the growing demand for data for training these models, developing efficient fine-tuning methods has become critical. This is particularly relevant for many domains and languages where obtaining large amounts of labeled training data is difficult or downright impossible. In such \textit{low-resource settings}, it becomes essential to effectively leverage and adapt PLMs while minimizing the need for extensive labeled data.


Data labeling is notoriously time-consuming and expensive, often hindering the development of sizable labeled datasets required for training high-performance models. \textit{Active learning} (AL) \cite{cohn-etal-1996-active, settles-2009-active} has emerged as a potential solution to this challenge. In contrast to \textit{passive learning}, in which the training set is sampled at random, AL encompasses a unique family of machine learning algorithms specifically designed to reduce labeling costs by reducing \textit{label complexity}, i.e., the number of labels required by an \textit{acquisition model} to achieve a certain level of performance \cite{dasgupta-2011-two}. With the advent of PLMs, AL research has pivoted towards investigating training regimes for PLMs, such as task-adaptive pre-training \cite[TAPT;][]{gururangan-etal-2020-dont}, that could be combined with AL to further reduce the label complexity.


While AL aims at directly minimizing the label complexity of learning, training efficiency can also be improved by
reducing the \textit{parameter complexity} of the model. This becomes more important as PLMs grow larger, and fine-tuning becomes increasingly challenging due to the sheer number of parameters involved.
To address this issue, adapters \cite{houlsby-etal-2019-parameter} have been introduced as compact modules that can be incorporated between the layers of PLMs. Adapters enable considerable parameter-sharing, facilitating parameter-efficient fine-tuning (PEFT) through modular learning \cite{pfeiffer-etal-2023-modular}. In this process, only the parameters of the adapters are updated during the tuning for a specific downstream task. Recent research \cite{he-etal-2021-effectiveness, li-liang-2021-prefix, karimi-mahabadi-etal-2021-parameter} has revealed that some PEFT methods outperform full fine-tuning (FFT) in low-resource settings, potentially due to better stability and a decreased risk of overfitting. In contrast, FFT has been shown to exhibit instability in scenarios with limited data.


Despite the promising results demonstrated by PEFT methods in low-resource settings, there is a striking gap in research on parameter-efficient training with respect to how PEFT interacts with AL.
Given that the majority of real-world AL scenarios involve a restricted amount of data, PEFT methods emerge as strong candidates for AL acquisition models. However, there has been no exploration of AL in conjunction with adapters. Investigating this uncharted territory can further advance our understanding of AL and reveal novel strategies for optimizing performance in low-resource settings.


In this paper, we present an empirical study on the behavior of PEFT in low-resource settings for text classification tasks. We analyze PEFT with and without AL and compare it against FFT. While our results confirm that PEFT exhibits superior performance in low-resource setups compared to FFT, we show that the improved performance with PEFT extends to AL scenarios in terms of performance gains over passive learning. Furthermore, we analyze the efficacy of TAPT in conjunction with AL and PEFT. We find that TAPT is beneficial in AL scenarios for both PEFT and fully fine-tuned models, thus representing a viable technique for improving performance in low-resource settings. Finally, aiming to illuminate why PEFT and TAPT improve AL performance in low-resource settings, we analyze the properties of PEFT and FFT via \textit{forgetting dynamics} \cite{toneva-etal-2019-empirical} and PLMs' instance-level representations. We find that AL methods choose fewer \textit{unforgettable} and more \textit{moderately forgettable} examples when combined with PEFT and TAPT, where forgetfulness indicates the model's tendency to learn and forget the gold label of a particular instance.
Compared to FFT, we observe that PEFT yields representations in the \textbf{early} and \textbf{middle} layers of a model that are more similar to the representations of the base PLM. We hypothesize that this property mitigates the issue of forgetting the knowledge obtained during pre-training when fine-tuning for downstream tasks.


In summary, we show that in AL low-resource settings for text classification, 
(1) PEFT yields greater performance improvements compared to FFT and
(2) TAPT enhances the overall classification performance of adapters and is well-suited for AL scenarios.
We also show that
(3) AL methods choose fewer unforgettable and more moderately forgettable examples with PEFT and
that (4) PEFT produces instance-level representations of early and middle layers that are more similar to the base PLM than FFT.
Our results uncover the intricacies of positive interactions between AL, PEFT, and TAPT, providing empirical justification for their combined use in low-resource settings.
\section{Related Work}
\label{sec:rw}

Our research involves combining AL with PLMs and investigating the use of PEFT techniques within the confines of low-resource settings.

\paragraph{AL with PLMs.}

Until recently, the conventional approach for integrating PLMs with AL involved performing full fine-tuning with a fixed number of training epochs and training the model from scratch in each AL step \cite{ein-dor-etal-2020-active, margatina-etal-2021-active, shelmanov-etal-2021-active, karamcheti-etal-2021-mind, schroder-etal-2022-revisiting}. However, studies by \citet{mosbach-etal-2021-stability} and \citet{zhang-etal-2021-revisiting} revealed that fine-tuning in low-resource setups is prone to instability, particularly when training for only a few epochs. This instability, often sensitive to weight initialization and data ordering \cite{dodge-etal-2020-fine}, presents a significant challenge for AL, which frequently operates in low-resource settings. Recent research has looked into the impact of PLM training regimes on AL performance \cite{griesshaber-etal-2020-fine, yuan-etal-2020-cold, yu-etal-2022-actune}, suggesting that the choice of training regime is more critical than the choice of the AL method. Notably, TAPT has proven particularly effective in enhancing AL performance \cite{margatina-etal-2022-importance, jukic-snajder-2023-smooth}.

\paragraph{Adapters in low-resource settings.}

Research on adapters in low-resource settings has primarily focused on areas such as cross-lingual transfer for low-resource languages \cite{ansell-etal-2021-mad-g, lee-etal-2022-fad, parovic-etal-2022-bad}, where the emphasis lies on exploring diverse methods of fusing adapters. In monolingual settings with scarce data, adapters have been found to outperform full fine-tuning \cite{li-liang-2021-prefix, mao-etal-2022-unipelt}. A study by \citet{he-etal-2021-effectiveness} demonstrated that adapter-based tuning exhibits enhanced stability and generalization capabilities by virtue of being less sensitive to learning rates than traditional fine-tuning methods. While incorporating task adaptation techniques, such as TAPT, has been shown to match or even improve performance over FFT in low-resource setups,
\citet{kim-etal-2021-revisiting} noted an interesting caveat: the benefits of integrating TAPT with adapters tend to taper off as the amount of data increases.

Despite the established effectiveness of adapters in setups with limited resources, their integration into AL frameworks --- which frequently face analogous resource constraints --- remains an untapped area of research. This gap is particularly notable given that AL's iterative learning process could significantly benefit from adapters' parameter efficiency and transferability, especially in scenarios where data scarcity or labeling costs are primary concerns.

\section{Preliminaries}
\label{sec:prelims}

We now describe the experimental setup, providing details on the datasets as well as the PEFT and AL methods used in our study.

\subsection{Datasets}

We employ four single-text classification tasks commonly used for AL evaluation: (1) the subjectivity dataset \cite[\textbf{\subj};][]{pang-lee-2004-sentimental}, designed to assess the subjectivity of a given text; (2) the question type classification dataset \cite[\textbf{\trec};][]{li-roth-2002-learning}, designed for categorizing questions according to their types; (3) the Stanford Sentiment Treebank \cite[\textbf{\sst};][]{socher-etal-2013-parsing}, which focuses on sentiment analysis; (4) AG's news classification dataset \cite[\textbf{\agn};][]{zhang-etal-2015-character}, which classifies news articles into different categories. We provide the dataset statistics in the appendix for further reference (cf.~Appendix \Cref{tab:dataset-stats}).

\subsection{PEFT methods}

We consider four prototypical PEFT techniques:

\begin{description}
\item[\textbf{\adapter}] incorporates trainable bottleneck layers after both the multi-head attention and feed-forward block in each Transformer layer \cite{houlsby-etal-2019-parameter};

\item[\textbf{\pt}] adds new parameters in the multi-head attention blocks within each Transformer layer \cite{li-liang-2021-prefix};

\item[\textbf{\lora}] (\textbf{Lo}w-\textbf{r}ank \textbf{a}daptation) represents an additive method that incorporates trainable low-rank decomposition matrices into the layers of a pre-trained model \cite{hu-etal-2022-lora};

\item[\textbf{\uni}] combines multiple PEFT approaches, namely LoRA, Prefix-tuning, and Adapter, in a single unified setup \cite{mao-etal-2022-unipelt}. Each constituent is a submodule, and UniPELT employs gating mechanisms to activate them effectively.
\end{description}


All of the above PEFT methods fall under the category of lightweight fine-tuning. While prefix-tuning does not technically qualify as an adapter, \citet{he-etal-2022-towards} demonstrated that it shares formal similarities with adapters, with prefix-tuning performing weighted addition and an adapter employing unweighted addition. We refer to all four considered methods as adapters for terminological simplicity. We use BERT \cite{devlin-etal-2019-bert} as the base PLM for every adapter. Additionally, we adhere to the hyperparameter settings for each adapter as recommended in the respective papers that introduced them (cf.~\Cref{app:ada} for details).

\subsection{AL methods}
Our study considers five sampling strategies, including \textbf{random selection} (\textbf{\rnd{}}) as a passive learning baseline. The other four strategies are AL methods originating from different families, chosen for their robustness (ability to perform well across various tasks) and widespread usage in the field:

\begin{description}
    
\item[\textbf{Maximum entropy}] \cite[\textbf{\ent{}};][]{lewis-gale-1994-sequential} comes from the family of \textit{uncertainty} strategies. The method queries instances where the model is least certain based on the maximum entropy criterion of the prediction output;

\item[\textbf{Monte Carlo dropout}] \cite[\textbf{\mc{}};][]{gal-ghahramani-2016-dropout} resembles \ent{} but utilizes the stochasticity of forward passes with dropout layers \cite{srivastava-etal-dropout} to estimate the entropy for a given instance;

\item[\textbf{Core-set}] \cite[\textbf{\cs{}};][]{sener-savarese-2018-active} encourages instance diversity by using the learned representations of the acquisition model. This method aims to minimize the distance between an example in the unlabeled set and its closest counterpart in the labeled subset;

\item[\textbf{Discriminative active learning}] \cite[\textbf{\dal{}};][]{gissin-shwartz-2019-discriminative} frames AL as a binary classification of instances into those that are labeled and those that are not, with the objective of making the labeled and unlabeled sets indistinguishable.

\end{description}

\subsection{Experimental setup}
In AL runs, we select 50 new examples in each step of each AL experiment, using $100$ examples for the warm start (randomly sampled labeled data to initiate the model). To probe different PEFT approaches with and without AL in low-resource settings, we establish a labeling budget limit of $1,000$ instances. To sidestep the need for a validation set in our experiments, which is typically unavailable in real-world AL scenarios, we adopt the Besov early stopping \cite{jukic-snajder-2023-smooth}. This method utilizes the smoothness of Transformer layers to decide at which epoch to stop training.

In the case of TAPT, we pre-train the base model on a masked language modeling task using unlabeled training data. For adapters, we only update the injected parameters while keeping the remaining parameters of the base model frozen. This approach aligns with the primary function of adapters, which is to utilize a common base model across diverse tasks. For every setting, we perform five runs using different random seeds. We report the average $F_1$ score at each sampling step (with and without AL for FFT and PEFT) to show the corresponding learning curve averaged over five runs. We provide details on training and hyperparameters in \Cref{app:hyper}.

\subsection{Evaluation}
\label{sec:prelim-eval}

To evaluate the overall performance of an AL method, we employ the area under the performance curve (\auc{}). In each individual AL step with a specific quantity of labeled examples, we measure the classification performance in terms of the $F_1$ score. The overall \auc{} is calculated using the $F_1$ scores obtained at each step. We advocate for using \auc{} alongside the AL curves, as \auc{} serves as a suitable approximation of AL feasibility through a summary numeric score, as recommended in recent AL literature \cite{schroder-etal-2022-revisiting, jukic-snajder-2023-smooth}.

As our experiments involve different training regimes, we compare each AL sampling strategy $S_{\mathrm{AL}}$ to passive learning $S_{\mathrm{PL}}$ within the same training regime to isolate the effects of AL.
The primary objective of AL is to improve label efficiency over passive learning. 
To test whether AL is successful, we calculate the \textbf{r}elative \textbf{i}mprovement over \textbf{p}assive \textbf{l}earning (RIPL), which we define as follows:
\[ \text{RIPL}(S_\mathrm{AL}, S_\mathrm{PL}) = \frac{\text{\auc}(S_\mathrm{AL}) - \text{\auc}(S_\mathrm{PL})}{1 - \text{\auc}(S_\mathrm{PL})}  \]
Intuitively, RIPL estimates the proportion of maximum possible improvement achievable by a given AL method compared to the passive learning baseline. A score of 1 indicates the maximum theoretical improvement, which would be tantamount to attaining an $F_1$ score of 1 in the initial sampling step and sustaining that score throughout all steps. Conversely, a negative score indicates that the AL method performs worse than passive learning.

\section{Experiments}
\label{sec:exp}

In this section, we first examine the performance of PEFT methods in comparison to FFT with passive learning and then proceed to analyze the application of PEFT in AL settings.

\subsection{PEFT vs.~FFT}


Previous research on the use of adapters in low-resource settings \cite{li-liang-2021-prefix, mao-etal-2022-unipelt, he-etal-2021-effectiveness} has demonstrated that adapters perform comparable to, and sometimes even better than FFT. However, these findings were based on comparing FFT to a single adapter variant on a full dataset or evaluating the performance at only a few discrete points.

In the first part of our experiments, we build upon these findings by conducting a more nuanced analysis. We generate detailed learning curves that facilitate the comparison of multiple adapters with FFT under the passive learning setup. Our comparison, summarized by the AUC metric in \Cref{tab:pl}, reveals that \uni{} and \pt{} consistently outperform FFT with a significant difference across all datasets used in our study. Conversely, the performance of \adapter{} and \lora{} is mostly comparable to FFT, although there are cases where they either outperform or underperform FFT. In cases in which \adapter{} and \lora{} perform better than FFT with significant differences, the degree of improvement is smaller than what is observed with \uni{} and \pt{}.

\begin{figure*}
    \centering
    \begin{subfigure}[b]{0.245\linewidth}
        \includegraphics[width=\linewidth]{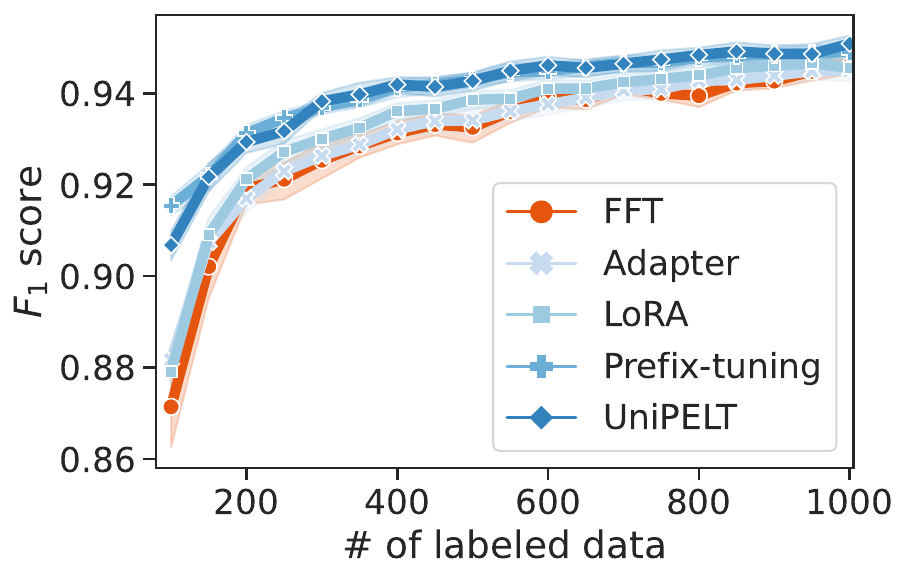}
        \caption{\subj}
        \label{fig:fft-vs-peft-subj}
    \end{subfigure}
    \begin{subfigure}[b]{0.245\linewidth}
        \includegraphics[width=\linewidth]{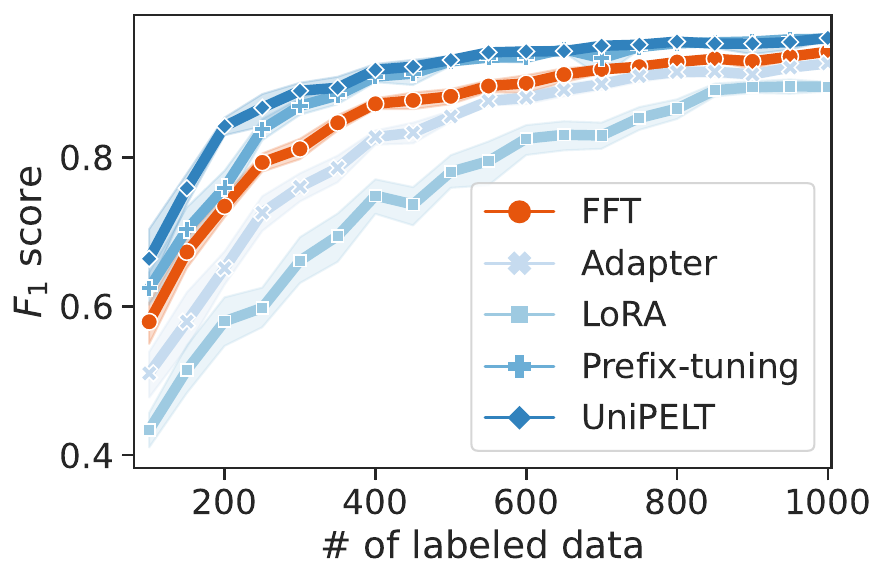}
        \caption{\trec}
        \label{fig:fft-vs-peft-trec}
    \end{subfigure}
    \begin{subfigure}[b]{0.245\linewidth}
        \includegraphics[width=\linewidth]{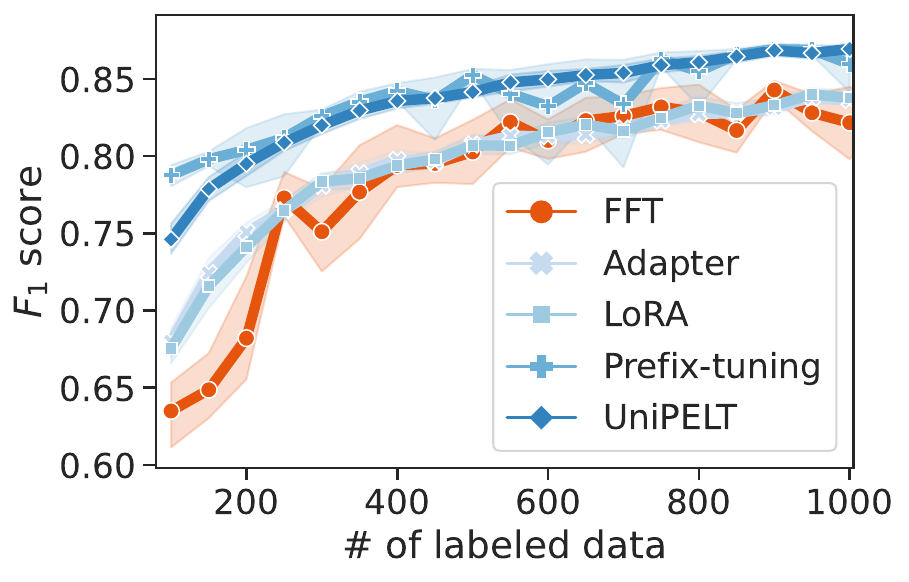}
        \caption{\sst}
        \label{fig:fft-vs-peft-sst}
    \end{subfigure}
    \begin{subfigure}[b]{0.245\linewidth}
        \includegraphics[width=\linewidth]{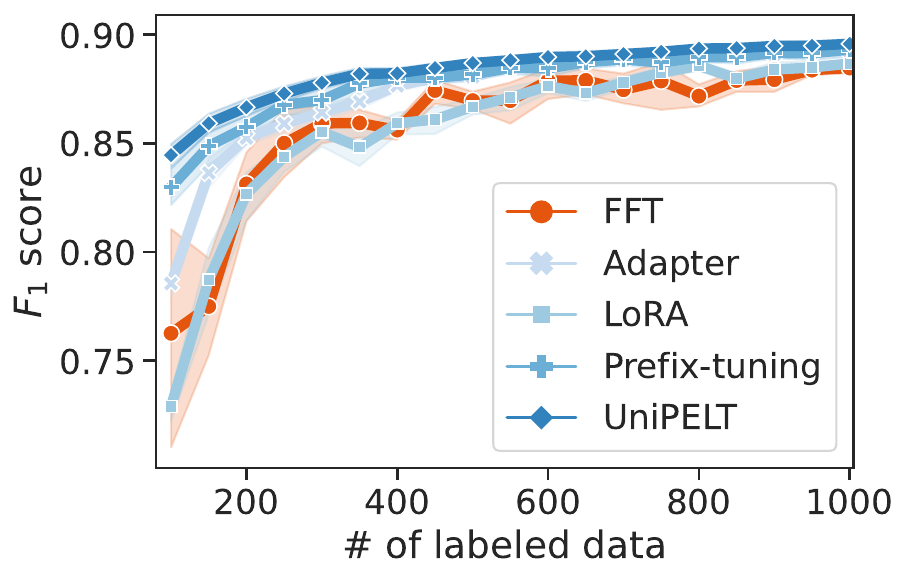}
        \caption{\agn}
        \label{fig:fft-vs-peft-agn}
    \end{subfigure}
\caption{Learning curves under the passive learning setup with different PEFT methods and FFT. The results are averaged over five runs. The shaded bands denote the standard deviation. Best viewed on a computer screen.}
\label{fig:fft-vs-peft}
\end{figure*}

Next, we look into how the models' performance changes as the training set increases. To that end, we show the corresponding learning curves for adapters and FFT in \Cref{fig:fft-vs-peft}. The performance disparities between adapters and FFT become more apparent under conditions of extreme data scarcity ($100$--$300$ labeled instances). Notably, the greatest differences in performance occur at the initial step (only $100$ labels). This highlights the promise of adapter-based methods in low-resource settings, particularly for \pt{} and \uni{}.

\begin{table}
\centering
\small
\begin{tabular}{llrrrr}
\toprule
& & \textsc{subj} & \textsc{trec} & \textsc{sst} & \textsc{agn} \\
\midrule
\multirow{4}{*}{\rotatebox[origin=c]{90}{adapters}}
& Adapter & $.926$ & $.804$ & $.800$\nospacetext{$^\dagger$} & $.871$\nospacetext{$^\dagger$}\\
& \lora{} & $.929$ & $.750$\nospacetext{$^\dagger$} & $.798$\nospacetext{$^\dagger$} & $.860$ \\
& \pt{} & $\textbf{.936}$\nospacetext{$^\dagger$} & $.847$\nospacetext{$^\dagger$} & $\textbf{.847}$\nospacetext{$^\dagger$} & $\textbf{.875}$\nospacetext{$^\dagger$} \\
& \uni{} & $.934$\nospacetext{$^\dagger$} & $\textbf{.877}$\nospacetext{$^\dagger$} & $.836$\nospacetext{$^\dagger$} & $\textbf{.875}$\nospacetext{$^\dagger$}\\
\midrule
& FFT & $.928$ & $.810$ & $.787$ & $.860$\\
\bottomrule
\end{tabular}
\caption{The performance of adapters and FFT in a passive learning setup in terms of the AUC metric (based on $F_1$ score) averaged over five runs. Numbers in \textbf{bold} represent the best-performing variant for a particular dataset. The ``$\dagger$'' symbol indicates when the mean \auc{} of an adapter is significantly different from the corresponding mean \auc{} of FFT ($p<.05$ using a two-sided Man-Whitney U test adjusted for family-wise error rate with the Holm-Bonferroni method).}
\label{tab:pl}
\end{table}

\subsection{PEFT with AL}

\begin{table*}[htb!]
\centering
\small
\begin{tabular}{lr|rrrr|rrrr}
\toprule
& & 
\multicolumn{4}{c}{without TAPT} & \multicolumn{4}{c}{with TAPT} \\
\cmidrule(lr){3-6} \cmidrule(lr){7-10} 
 & & \ent{} & \mc{} & \cs{} & \dal{} & \ent{} & \mc{} & \cs{} & \dal{} \\
\midrule
\multirow{5}{*}{\rotatebox[origin=c]{90}{\subj}}
& FFT & $.050$ & $.059$ & $.061$ & $.077$ & $.140$ & $.140$ & $.142$ & $.126$\\
& \adapter{} & $.112$ & $.102$ & $.100$ & $.092$ & $.137$ & $.151$ & $.111$ & $.067$\\
& \lora{} & $.127$ & $.115$ & $.091$ & $.081$ & $.165$ & $.160$ & $.122$ & $.100$\\
& \pt & $.095$ & $.110$ & $.106$ & $.111$ & $\textbf{.186}$ & $.181$ & $.170$ & $.151$\\
& \uni{} & $.129$ & $\textbf{.153}$ & $.131$ & $.128$ & $.159$ & $.167$ & $.163$ & $.157$\\
\midrule
\multirow{5}{*}{\rotatebox[origin=c]{90}{\trec}}
& FFT & $.011$ & $.022$ & $.038$ & $.034$ & $.162$ & $.180$ & $.141$ & $.159$\\
& \adapter{} & $.027$ & $.069$ & $.137$ & $.084$ & $.124$ & $.146$ & $.079$ & $.154$\\
& \lora{} &$.098$ & $.065$ & $.048$ & $.007$ & $.254$ & $.237$ & $.243$ & $.074$\\
& \pt{} & $.093$ & $.105$ & $.068$ & $.093$ & $.246$ & $.227$ & $.205$ & $.241$\\
& \uni{} & $.138$ & $.165$ & $.082$ & $\textbf{.200}$ & $.302$ & $\textbf{.334}$ & $.276$ & $.236$\\
\midrule
\multirow{5}{*}{\rotatebox[origin=c]{90}{\sst}}
& FFT & $.002$ & $.011$ & $-.039$ & $.004$ & $.080$ & $.079$ & $.075$ & $.070$\\
& \adapter{} & $.015$ & $.048$ & $.025$ & $.002$ & $.035$ & $.034$ & $.028$ & $.008$\\
& \lora{} & $.001$ & $.007$ & $.064$ & $.031$ & $.036$ & $.022$ & $.032$ & $.014$\\
& \pt{} & $.049$ & $.060$ & $\textbf{.114}$ & $.031$ & $\textbf{.152}$ & $.143$ & $.137$ & $.126$\\
& \uni{} & $.037$ & $.043$ & $.040$ & $.008$ & $.082$ & $.101$ & $.083$ & $.080$\\
\midrule
\multirow{5}{*}{\rotatebox[origin=c]{90}{\agn}}
& FFT & $.014$ & $.032$ & $.007$ & $.092$ & $.134$ & $.021$ & $.089$ & $.017$ \\
& \adapter{} & $.074$ & $.046$ & $.015$ & $.062$ & $.115$ & $.089$ & $.077$ & $.080$\\
& \lora{} & $.020$ & $.025$ & $.067$ & $.016$ & $.028$ & $.102$ & $.071$ & $.023$\\
& \pt{} & $.054$ & $.023$ & $.040$ & $.033$ & $.035$ & $.143$ & $.098$ & $.092$\\
& \uni{} & $.074$ & $\textbf{.096}$ & $.089$ & $.095$ & $\textbf{.185}$ & $.151$ & $.112$ & $.081$  \\
\bottomrule
\end{tabular}
\caption{Improvement over passive learning in terms of the RIPL metric for four AL methods considered (\ent{}, \mc{}, \cs{}, and \dal{}) and for all combinations of adapters and datasets considered, shown separately without TAPT and with TAPT. Positive values indicate improvement over passive learning, while negative values indicate performance drops compared to passive learning. Values in \textbf{bold} denote the best result for a particular dataset across different adapters and AL methods within the same regime (with or without TAPT).}
\label{tab:aucs}
\end{table*}


Motivated by our initial findings on using PEFT under the passive learning setup, where PEFT exhibited promising properties in low-resource settings, we further explore the behavior of adapters in AL scenarios.
We evaluate individual PEFT methods in AL scenarios with and without using TAPT in terms of gains over random sampling (passive learning) using the RIPL metric described in \Cref{sec:prelim-eval}. \Cref{tab:aucs} shows the results for different combinations of AL methods and adapters, evaluated through the RIPL metric. We complement these results with absolute values in terms of AUC (cf.~Appendix \Cref{tab:aucs-app}). For FFT without TAPT, \dal{} achieved the highest RIPL score on two datasets, while \cs{} and \mc{} topped the chart on one dataset each. When we incorporated TAPT, \ent{} yielded the best results on three out of four datasets, with \cs{} leading on one. Looking at adapters, the most successful AL methods without TAPT vary, depending on the specific adapter and dataset in question. Interestingly, when TAPT is applied, the best results for all adapters are obtained either by \ent{} or \mc{}. We speculate this could be attributed to solid compatibility between entropy-based methods and TAPT when adapters are employed.

\begin{figure}
\centering
\includegraphics[width=\linewidth, height=4cm]{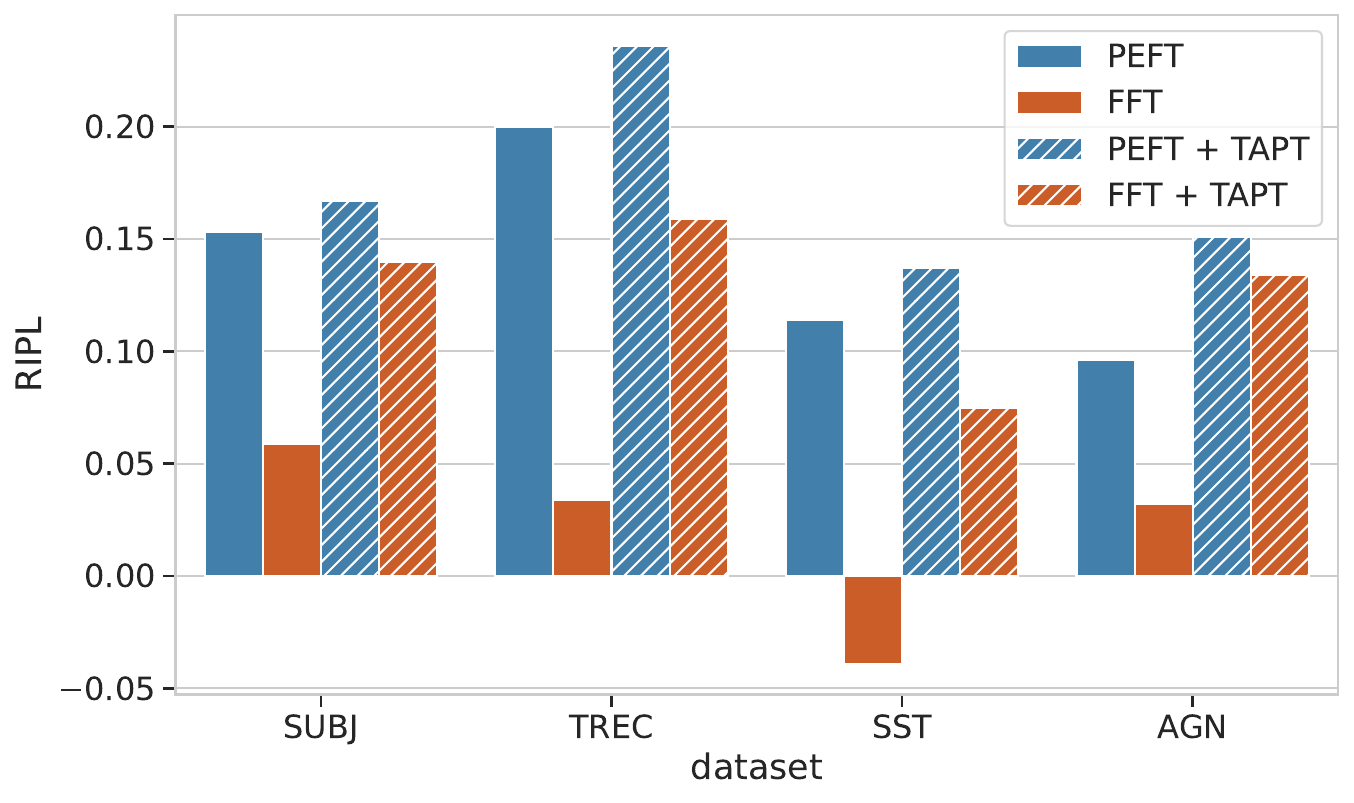}
\caption{Comparison of best-performing adapters and FFT from \Cref{tab:aucs} and their corresponding version with TAPT applied.}
\label{fig:tapt-bar}
\end{figure}

Furthermore, we observe that without TAPT, adapters achieve larger gains over FFT. However, when TAPT is applied, FFT becomes comparable to PEFT, although \pt{} and \uni{} still yield the greatest improvements, depending on the dataset and AL method used. In \Cref{fig:tapt-bar}, we select the adapters that achieved the best improvement according to \Cref{tab:aucs} without TAPT and show their RIPL value compared against FFT as well as their corresponding version when TAPT is applied. We conjecture that TAPT reduces the performance gap between adapters and FFT by inducing FFT to emulate PEFT in aspects such as training dynamics and representation space --- a hypothesis we explore in more detail in \Cref{sec:analysis}.

\begin{figure*}[h!]
\centering
\includegraphics[width=\linewidth, height=0.65\linewidth]{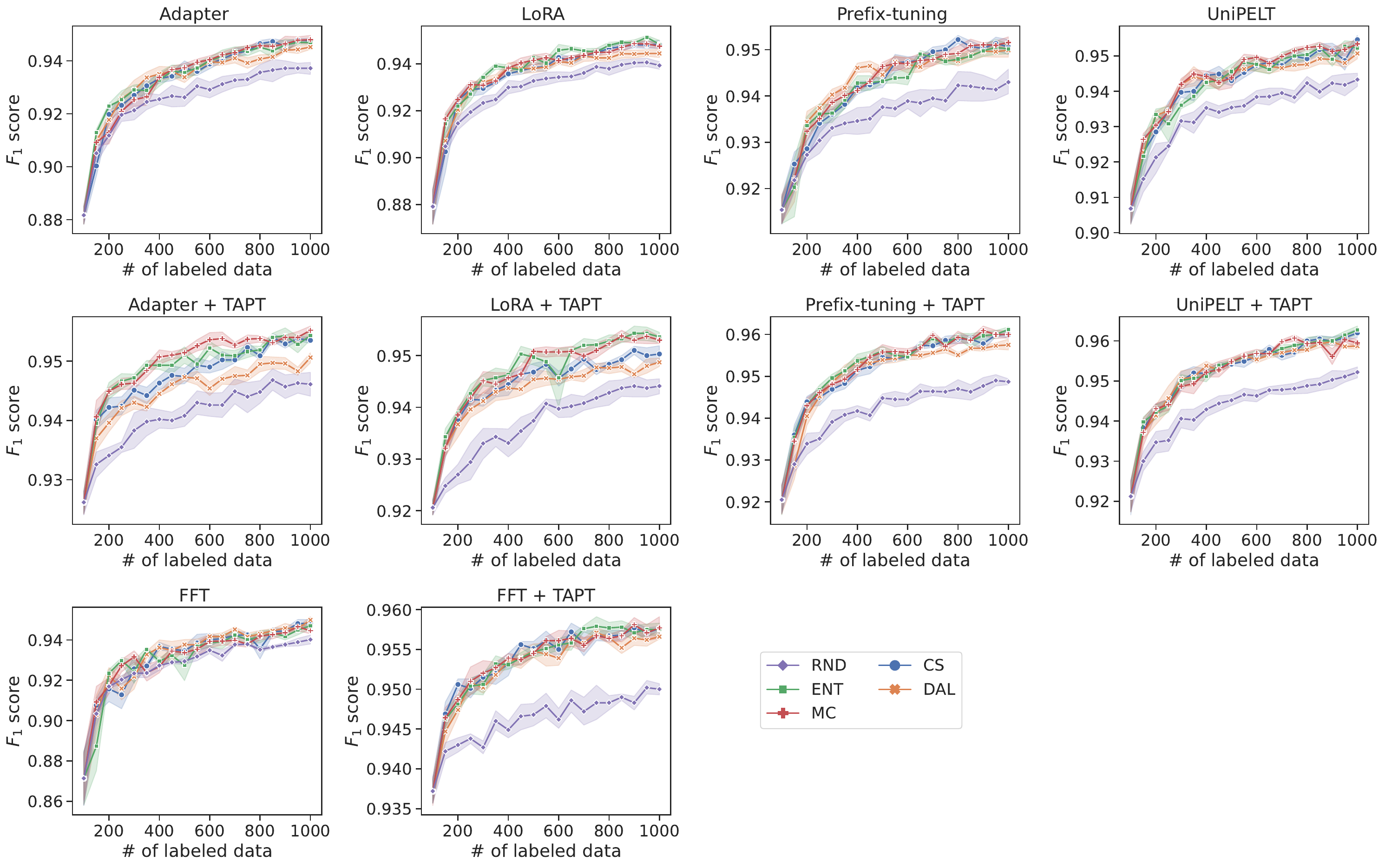}
\caption{AL learning curves compared with random sampling on the \subj{} dataset. The first and the second rows show learning curves for adapters without and with TAPT, respectively. The third row shows learning curves for FFT, without and with TAPT. The results are averaged over five runs, and the shaded bands denote the standard deviation. Best viewed on a computer screen.}
\label{fig:subj}
\end{figure*}

We further investigate the behavior of adapters with AL throughout the individual steps. \Cref{fig:subj} shows the learning curves for corresponding adapter models with and without applying TAPT. Due to space constraints, we show the learning curves only for the \subj{} dataset, as similar trends occur for other datasets. Without TAPT, the performance of adapters is largely independent of the specific AL method used, where \pt{} and \uni{} consistently outperform \adapter{} and \lora{} across all AL steps. With TAPT, the differences between AL and random sampling are more pronounced starting from the early steps, typically already with $200$ instances. In contrast, the gap becomes more apparent only with $500$ or more instances when TAPT is not employed.

\section{Analysis}
\label{sec:analysis}

\begin{figure*}[h]
    \centering
    \begin{subfigure}[b]{0.325\linewidth}
        \includegraphics[width=\linewidth, height=3cm]{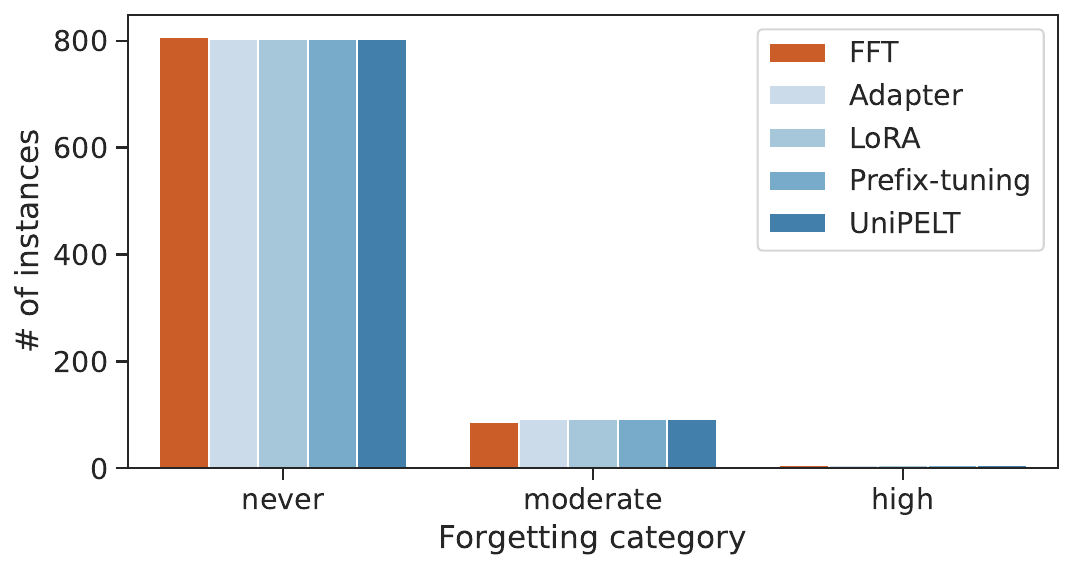}
        \caption{\subj{}; \rnd{}}
        \label{fig:forget_subj_rnd}
    \end{subfigure}
    \begin{subfigure}[b]{0.325\linewidth}
        \includegraphics[width=\linewidth, height=3cm]{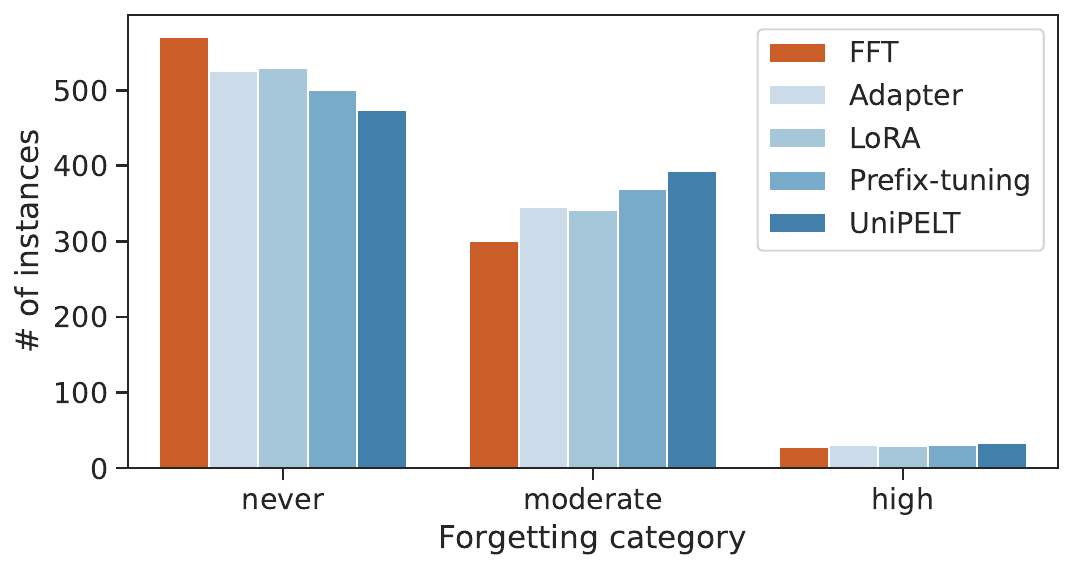}
        \caption{\subj{}; \mc{}}
        \label{fig:forget_subj_mc}
    \end{subfigure}
    \begin{subfigure}[b]{0.325\linewidth}
        \includegraphics[width=\linewidth, height=3cm]{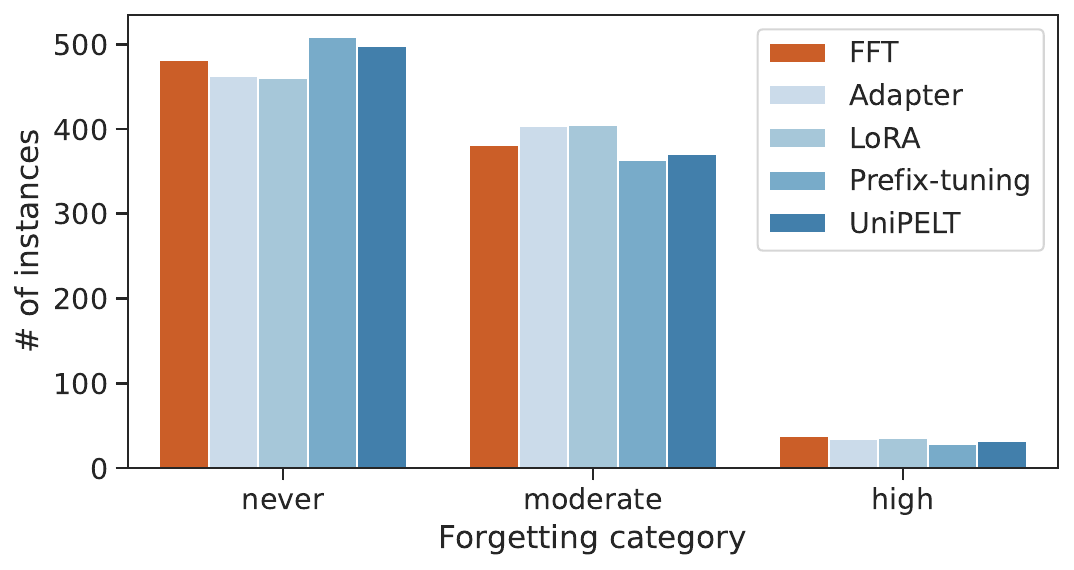}
        \caption{\subj{}; \mc{} + TAPT}
        \label{fig:forget_subj_mc_tapt}
    \end{subfigure}
    \begin{subfigure}[b]{0.325\linewidth}
        \includegraphics[width=\linewidth, height=3cm]{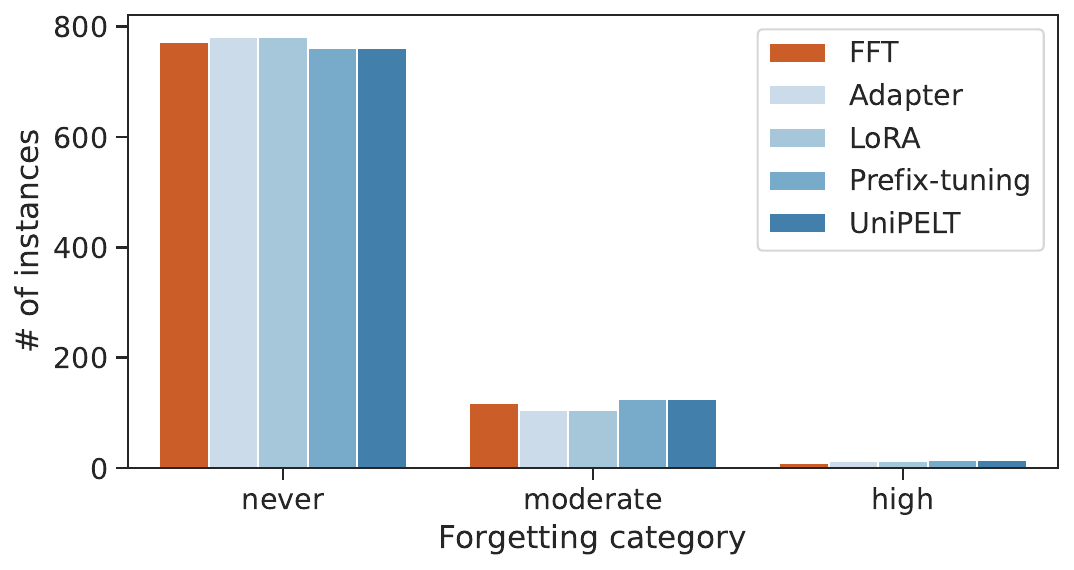}
        \caption{\trec{}; \rnd{}}
        \label{fig:forget_trec_rnd}
    \end{subfigure}
    \begin{subfigure}[b]{0.325\linewidth}
        \includegraphics[width=\linewidth, height=3cm]{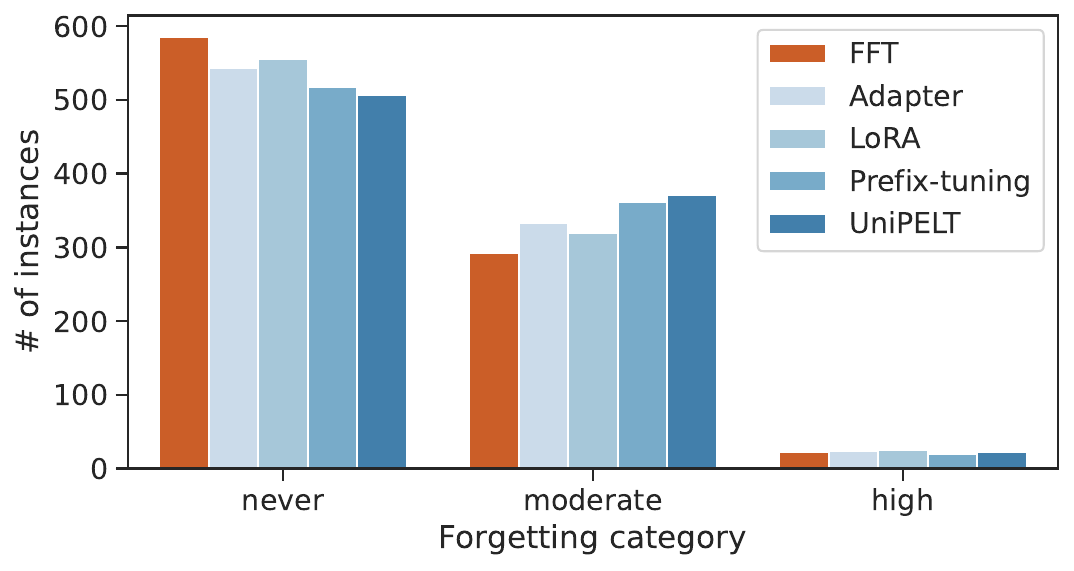}
        \caption{\trec{}; \mc{}}
        \label{fig:forget_trec_mc}
    \end{subfigure}
    \begin{subfigure}[b]{0.325\linewidth}
        \includegraphics[width=\linewidth, height=3cm]{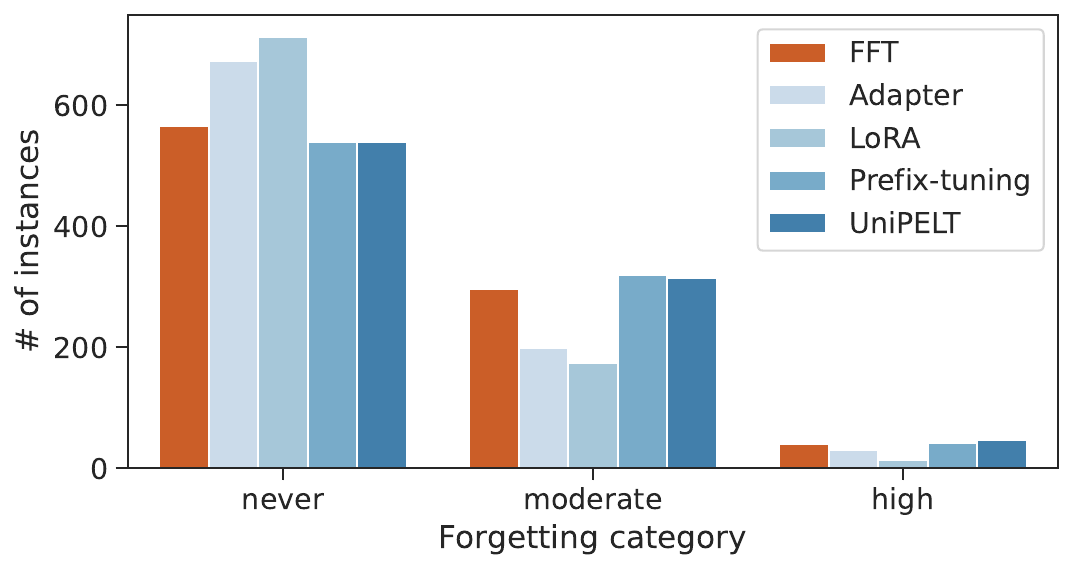}
        \caption{\trec{}; \mc{} + TAPT}
        \label{fig:forget_trec_mc_tapt}
    \end{subfigure}
\caption{Forgetting dynamics for random sampling (passive learning) and AL with \mc{} without and with TAPT on \subj{} and \trec{}. The x-axis shows the number of instances in each of the forgetting categories: the ``never'' category representing \textbf{unforgettable} instances, \textbf{moderately} forgettable instances, and \textbf{highly} forgettable instances.}
\label{fig:forget}
\end{figure*}

\begin{figure*}[h]
    \centering
    \begin{subfigure}[b]{.195\linewidth}
        \includegraphics[width=\linewidth]{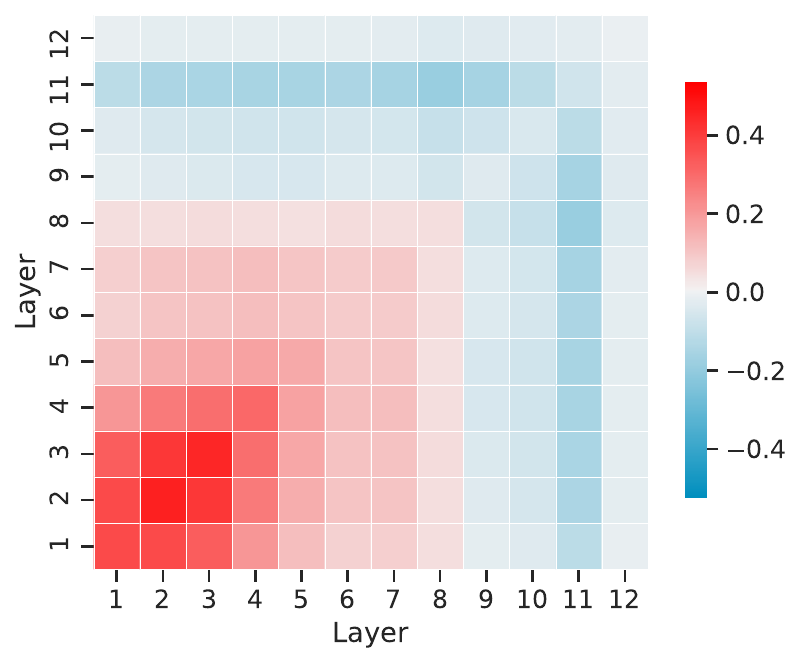}
        \caption{\ent{}}
        \label{fig:repr_ent}
    \end{subfigure}
    \begin{subfigure}[b]{.195\linewidth}
        \includegraphics[width=\linewidth]{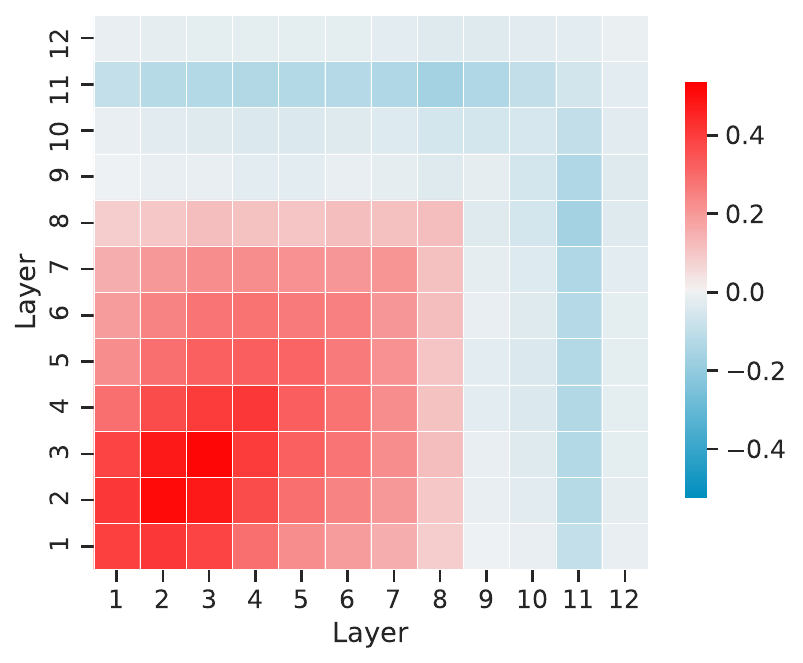}
        \caption{\mc{}}
        \label{fig:repr_mc}
    \end{subfigure}
    \begin{subfigure}[b]{0.195\linewidth}
        \includegraphics[width=\linewidth]{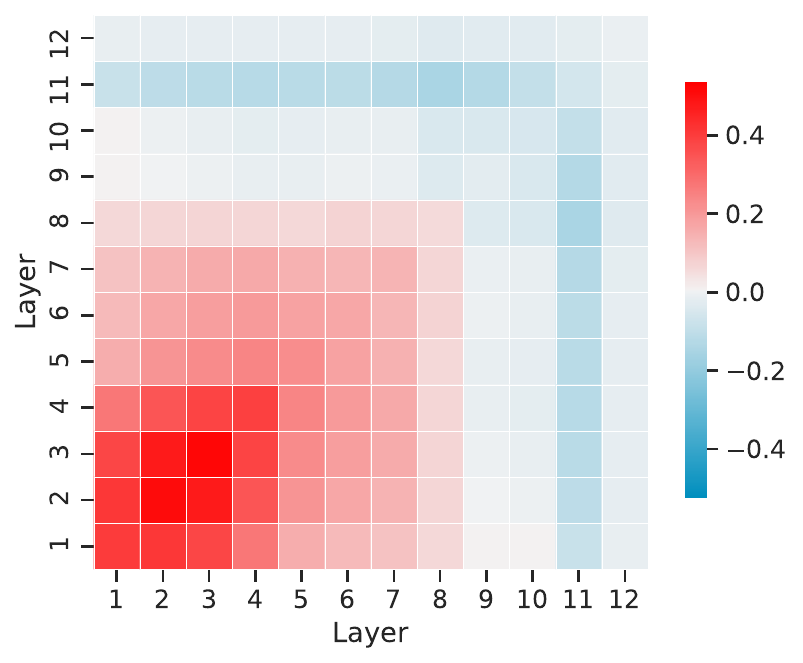}
        \caption{\cs{}}
        \label{fig:repr_cs}
    \end{subfigure}
    \begin{subfigure}[b]{0.195\linewidth}
        \includegraphics[width=\linewidth]{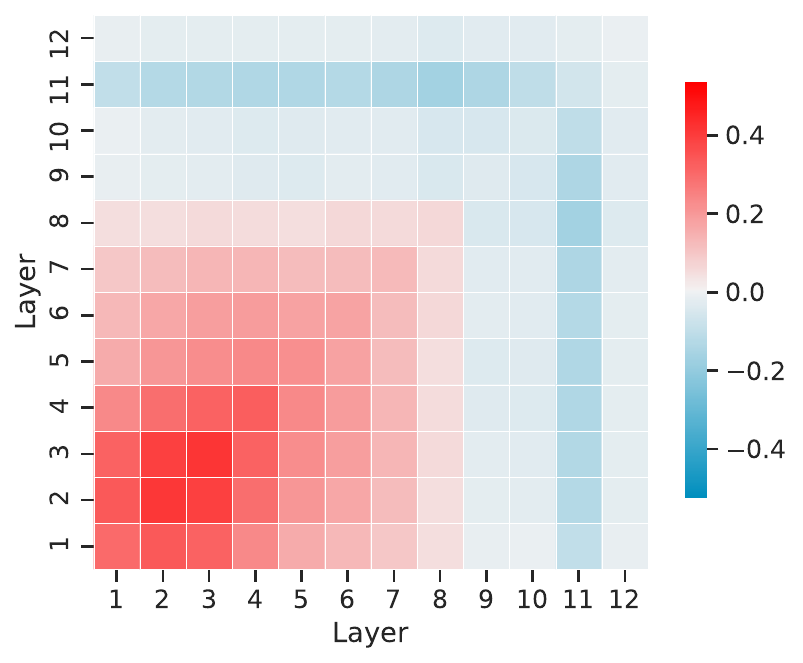}
        \caption{\dal{}}
        \label{fig:repr_dal}
    \end{subfigure}
    \begin{subfigure}[b]{0.195\linewidth}
        \includegraphics[width=\linewidth]{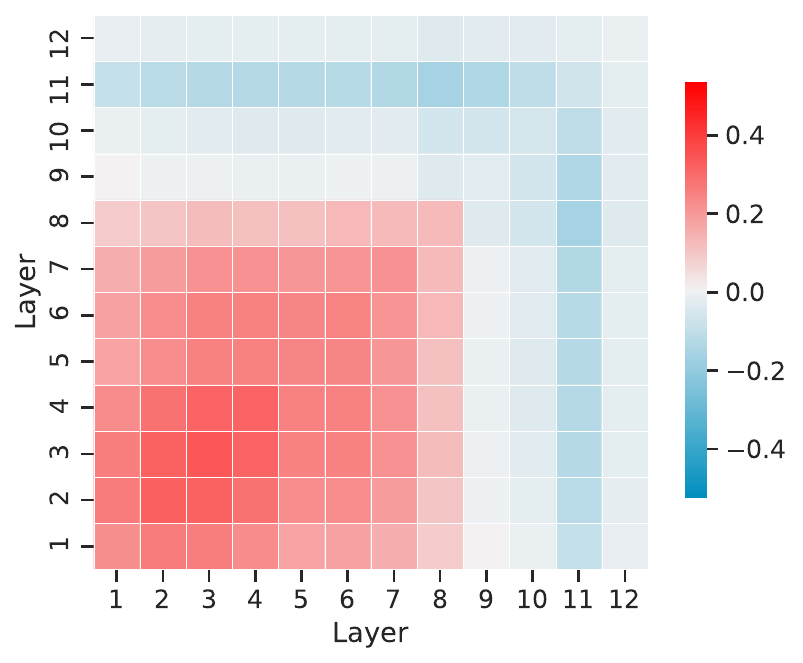}
        \caption{\rnd{}}
        \label{fig:repr_rnd}
    \end{subfigure}
\caption{Layerwise difference in representation similarity for the \uni{} adapter and the FFT model on \subj{}. We observed similar patterns in other adapters and datasets we used (cf.~\Cref{app:exp}). Warm colors (positive values) illustrate layer pairs that demonstrate higher similarity to the base model with the adapter than with FFT. Conversely, cool colors (negative values) represent layer pairs that are more similar to the base model when using the FFT model. Best viewed on a computer screen.}
\label{fig:repr_diff}
\end{figure*}

In \Cref{sec:exp}, we have demonstrated that PEFT exhibits larger gains than FFT when combined with AL in low-resource settings, which is also accompanied by superior performance with passive leaning. To better understand why PEFT displays superior behavior with limited data, we now examine two specific properties of adapters and FFT models. First, we analyze the influence of TAPT on the forgetting dynamics during training. We continue with example-level representation analysis, where we investigate the representation similarity of PEFT and FFT to their respective base models.

\subsection{Forgetting dynamics}

We employ forgetting dynamics to compare PEFT and FFT's learning stability and their impact on AL data selection. The underlying hypothesis is that having fewer forgetting events in adapters would indicate a more stable and effective learning process.
In utilizing forgetting dynamics, we draw upon the study by \citet{toneva-etal-2019-empirical}, focusing on the occurrence of \textit{forgetting events} --- cases where a specific training example transitions from correct to incorrect classification over the course of multiple learning epochs. More specifically, we divide the instances into three categories: (1) \textbf{unforgettable} instances, i.e., the ones that have never experienced a forgetting event during training, (2) instances that have encountered one or two forgetting events, labeled as \textbf{moderately} forgettable, and (3) instances subjected to three or more forgetting events, referred to as \textbf{highly} forgettable instances. As pointed out in the original study, moderately forgettable, \textit{ambiguous} instances are more valuable for the learning model than unforgettable, \textit{easy} instances. However, it is worth noting that AL is often hindered by too \textit{hard} or \textit{impossible-to-learn} examples \cite{karamcheti-etal-2021-mind}, which roughly correspond to the highly forgettable examples.

\Cref{fig:forget} shows the distribution of instances across the three categories of forgetting events for \subj{} and \trec{} datasets. We focus on these two datasets as examples of a simple binary classification task and a more complex multi-class classification task, respectively. Specifically, we compare \rnd{} with \mc{}, which achieves consistent performance improvements across all datasets. Our findings suggest that FFT tends to select a higher number of unforgettable instances and fewer moderately forgettable instances when compared to adapters. Interestingly, the adapters that perform best --- \pt{} and \uni{} --- appear to favor moderately forgettable instances. However, when TAPT is applied, the discrepancies in forgetting profiles between FFT and the top two adapters, \pt{} and \uni{}, seem to diminish. In contrast, TAPT amplifies the differences between FFT and the other two adapters, \lora{} and \adapter{}, which typically show smaller improvements than \pt{} and \uni{}. Given their superior AL performance, we hypothesize that the forgetting profiles of \pt{} and \uni{} are more favorable compared to other adapters. Moreover, FFT with TAPT approaches the performance of the superior adapters and simultaneously develops a forgetting profile similar to theirs.

\subsection{Representation analysis}

To bolster our findings, we explore the representations of adapters and FFT models. As suggested in previous research \cite{he-etal-2021-effectiveness, li-liang-2021-prefix, mao-etal-2022-unipelt}, adapters often display greater stability in terms of loss, especially in scenarios with limited resources. Our aim is to examine the stability of their representations and their relationship with overall AL performance.

We draw inspiration from research by \citet{stephenson-etal-2021-geometry} and \citet{baldock-etal-2021-deep}, which suggests that different layers of networks specialize in different features --- earlier layers tend to acquire more generalized knowledge, while the deeper layers are more focused on task-specific information. This leads us to a layerwise examination of similarity. To analyze the effect of PEFT and FFT on AL selection with respect to their layerwise similarity to the base model, we utilize centered kernel alignment (CKA) as a similarity measure between two sets of representations \cite{kornblith-etal-2019-similarity}. It has been shown that PEFT methods result in representations closer to the base model at the token level \cite{he-etal-2021-effectiveness}. We extend the analysis to example-level representation to explore the behavior of models with AL. We opt for CKA as it is designed to be invariant to invertible linear transformation and still can measure meaningful similarities between representations of higher dimensions than the number of data points. This stands in contrast to other metrics, which frequently falter when dealing with high-dimensional representations.

For a more direct comparison between PEFT and FFT, we analyze the differences between their respective similarities to their base models. Specifically, we compute the difference $\mathrm{CKA}(\textit{adapter}, \textit{base}) - \mathrm{CKA}(\textit{FFT}, \textit{base})$ for a specific adapter or FFT and their base models. We hypothesize that superior PEFT performance with AL compared to FFT will be accompanied by a more similar early layer representation to the base model in PEFT. \Cref{fig:repr_diff} visualizes the layerwise difference in similarity between the base model and the adapter model and between the base model and the FFT model. We find that PEFT representations are more similar to the base model in the \textbf{early} and \textbf{middle} layers when compared to FFT. This holds for all AL methods, with differences more pronounced than in passive learning. Specifically, up to the eighth layer, representations are much more similar in adapters than in FFT models. In the final four layers, the difference in CKA scores between the adapter and FFT model is close to zero. Interestingly, the penultimate layer is more similar in the FFT model with respect to the base model.

When fine-tuning on a downstream task, we believe that the increased stability of PEFT in earlier layers, relative to FFT, is instrumental in retaining the foundational knowledge from the PLM's pre-training phase. Conversely, PEFT exhibits more substantial transformations in the later, more task-specific layers. This ensures the preservation of essential pre-trained knowledge while allowing for task-relevant flexibility. We speculate that this strategic balance in PEFT influences its propensity to select moderately forgettable instances when combined with AL, contributing to its enhanced performance over FFT. These instances are neither too trivial to provide no learning value, nor are they too complex to risk misinterpretation, thereby enhancing the effectiveness of learning.

\section{Conclusion}

Our study has shed light on the advantages of parameter-efficient fine-tuning (PEFT) in low-resource settings, confirming its superiority over full fine-tuning (FFT) methods. Importantly, we have demonstrated that the integration of PEFT with active learning (AL) can offer substantial performance gains compared to passive learning, even in settings where labeled data is scarce. Furthermore, we highlighted the potential of task-adaptive pre-training (TAPT) to improve model performance further when used in conjunction with both PEFT and AL.
We found that AL methods, in combination with PEFT, tend to select fewer \textit{unforgettable} instances and more \textit{moderately forgettable} examples. We further found that PEFT maintains the integrity of \textbf{early} and \textbf{middle} layer representations similar to the base model. We conjecture that this property mitigates forgetting during downstream task fine-tuning. These insights inform us of a possible underpinning mechanism that contributes to PEFT's superior performance and stability in low-resource settings.
Overall, our work highlights the potential of PEFT and AL and establishes a foundation for developing increasingly efficient and cost-effective approaches for training models in low-resource settings.

\section*   {Limitations}
\label{sec:limitations}

While our study advances the understanding of PEFT and AL's interaction in low-resource settings and uncovers intriguing insights about the forgetting dynamics during fine-tuning, it has a number of limitations.

To begin with, we have focused on text classification tasks, which are but one aspect of the wide range of potential applications for PLMs. Different tasks such as question answering, translation, or summarization might exhibit different behaviors under the same conditions. Consequently, the observed advantages of PEFT in the context of AL might not necessarily translate to other NLP tasks.

Next, our results are limited to the specific PLMs, AL strategies, and PEFT methods we have examined in this study. While we have attempted to be comprehensive in our experiments, the outcomes might vary with different models, strategies, or methods. For example, the effectiveness of AL combined with PEFT might differ if other AL strategies are employed. Similarly, different types of adapter architectures could potentially lead to different results.

Although we found that PEFT methods produce instance-level representations of early and middle layers more similar to the base PLM than FFT, a comprehensive understanding of how and why this similarity leads to increased stability and performance in low-resource settings is still lacking. Our hypothesis about the role of early and middle layer stability in mitigating the issue of forgetting the knowledge obtained during pre-training needs further substantiation.

Finally, it is important to acknowledge the complexity and multifaceted nature of forgetting dynamics. While our investigation provides valuable insights about the interaction of forgetting with PEFT and TAPT in AL scenarios, a deeper understanding of the mechanisms of forgetting in the context of large PLMs is needed. Particularly, it would be interesting to investigate whether the balance between unforgettable and moderately forgettable instances selected by the AL methods changes as the size of the model or the amount of available data changes.

Future work should aim to address these limitations and further explore the mechanisms behind the promising results obtained with the combination of PEFT and AL. This will contribute to a more comprehensive understanding of the interaction between AL and PLMs, and help refine strategies for efficient fine-tuning in low-resource settings.


\bibliography{references/anthology, references/custom}

\begin{thebibliography}{44}
\expandafter\ifx\csname natexlab\endcsname\relax\def\natexlab#1{#1}\fi

\bibitem[{Ansell et~al.(2021)Ansell, Ponti, Pfeiffer, Ruder, Glava{\v{s}},
  Vuli{\'c}, and Korhonen}]{ansell-etal-2021-mad-g}
Alan Ansell, Edoardo~Maria Ponti, Jonas Pfeiffer, Sebastian Ruder, Goran
  Glava{\v{s}}, Ivan Vuli{\'c}, and Anna Korhonen. 2021.
\newblock \href {https://doi.org/10.18653/v1/2021.findings-emnlp.410}
  {{MAD}-{G}: {M}ultilingual adapter generation for efficient cross-lingual
  transfer}.
\newblock In \emph{Findings of the Association for Computational Linguistics:
  EMNLP 2021}, pages 4762--4781, Punta Cana, Dominican Republic. Association
  for Computational Linguistics.

\bibitem[{Baldock et~al.(2021)Baldock, Maennel, and
  Neyshabur}]{baldock-etal-2021-deep}
Robert Baldock, Hartmut Maennel, and Behnam Neyshabur. 2021.
\newblock \href
  {https://proceedings.neurips.cc/paper/2021/file/5a4b25aaed25c2ee1b74de72dc03c14e-Paper.pdf}
  {Deep learning through the lens of example difficulty}.
\newblock In \emph{Advances in Neural Information Processing Systems},
  volume~34, pages 10876--10889. Curran Associates, Inc.

\bibitem[{Cohn et~al.(1996)Cohn, Ghahramani, and
  Jordan}]{cohn-etal-1996-active}
David~A Cohn, Zoubin Ghahramani, and Michael~I Jordan. 1996.
\newblock Active learning with statistical models.
\newblock \emph{Journal of artificial intelligence research}, 4:129--145.

\bibitem[{Dasgupta(2011)}]{dasgupta-2011-two}
Sanjoy Dasgupta. 2011.
\newblock \href {https://doi.org/https://doi.org/10.1016/j.tcs.2010.12.054}
  {Two faces of active learning}.
\newblock \emph{Theoretical Computer Science}, 412(19):1767--1781.
\newblock Algorithmic Learning Theory (ALT 2009).

\bibitem[{Devlin et~al.(2019)Devlin, Chang, Lee, and
  Toutanova}]{devlin-etal-2019-bert}
Jacob Devlin, Ming-Wei Chang, Kenton Lee, and Kristina Toutanova. 2019.
\newblock \href {https://doi.org/10.18653/v1/N19-1423} {{BERT}: Pre-training of
  deep bidirectional transformers for language understanding}.
\newblock In \emph{Proceedings of the 2019 Conference of the North {A}merican
  Chapter of the Association for Computational Linguistics: Human Language
  Technologies, Volume 1 (Long and Short Papers)}, pages 4171--4186,
  Minneapolis, Minnesota. Association for Computational Linguistics.

\bibitem[{Dodge et~al.(2020)Dodge, Ilharco, Schwartz, Farhadi, Hajishirzi, and
  Smith}]{dodge-etal-2020-fine}
Jesse Dodge, Gabriel Ilharco, Roy Schwartz, Ali Farhadi, Hannaneh Hajishirzi,
  and Noah Smith. 2020.
\newblock Fine-tuning pretrained language models: Weight initializations, data
  orders, and early stopping.
\newblock \emph{arXiv preprint arXiv:2002.06305}.

\bibitem[{Ein-Dor et~al.(2020)Ein-Dor, Halfon, Gera, Shnarch, Dankin, Choshen,
  Danilevsky, Aharonov, Katz, and Slonim}]{ein-dor-etal-2020-active}
Liat Ein-Dor, Alon Halfon, Ariel Gera, Eyal Shnarch, Lena Dankin, Leshem
  Choshen, Marina Danilevsky, Ranit Aharonov, Yoav Katz, and Noam Slonim. 2020.
\newblock \href {https://doi.org/10.18653/v1/2020.emnlp-main.638} {{A}ctive
  {L}earning for {BERT}: {A}n {E}mpirical {S}tudy}.
\newblock In \emph{Proceedings of the 2020 Conference on Empirical Methods in
  Natural Language Processing (EMNLP)}, pages 7949--7962, Online. Association
  for Computational Linguistics.

\bibitem[{Gal and Ghahramani(2016)}]{gal-ghahramani-2016-dropout}
Yarin Gal and Zoubin Ghahramani. 2016.
\newblock \href {https://proceedings.mlr.press/v48/gal16.html} {Dropout as a
  bayesian approximation: Representing model uncertainty in deep learning}.
\newblock In \emph{Proceedings of The 33rd International Conference on Machine
  Learning}, volume~48 of \emph{Proceedings of Machine Learning Research},
  pages 1050--1059, New York, New York, USA. PMLR.

\bibitem[{Gissin and Shalev-Shwartz(2019)}]{gissin-shwartz-2019-discriminative}
Daniel Gissin and Shai Shalev-Shwartz. 2019.
\newblock Discriminative active learning.
\newblock \emph{arXiv preprint arXiv:1907.06347}.

\bibitem[{Grie{\ss}haber et~al.(2020)Grie{\ss}haber, Maucher, and
  Vu}]{griesshaber-etal-2020-fine}
Daniel Grie{\ss}haber, Johannes Maucher, and Ngoc~Thang Vu. 2020.
\newblock \href {https://doi.org/10.18653/v1/2020.coling-main.100} {Fine-tuning
  {BERT} for low-resource natural language understanding via active learning}.
\newblock In \emph{Proceedings of the 28th International Conference on
  Computational Linguistics}, pages 1158--1171, Barcelona, Spain (Online).
  International Committee on Computational Linguistics.

\bibitem[{Gururangan et~al.(2020)Gururangan, Marasovi{\'c}, Swayamdipta, Lo,
  Beltagy, Downey, and Smith}]{gururangan-etal-2020-dont}
Suchin Gururangan, Ana Marasovi{\'c}, Swabha Swayamdipta, Kyle Lo, Iz~Beltagy,
  Doug Downey, and Noah~A. Smith. 2020.
\newblock \href {https://doi.org/10.18653/v1/2020.acl-main.740} {Don{'}t stop
  pretraining: Adapt language models to domains and tasks}.
\newblock In \emph{Proceedings of the 58th Annual Meeting of the Association
  for Computational Linguistics}, pages 8342--8360, Online. Association for
  Computational Linguistics.

\bibitem[{He et~al.(2022)He, Zhou, Ma, Berg-Kirkpatrick, and
  Neubig}]{he-etal-2022-towards}
Junxian He, Chunting Zhou, Xuezhe Ma, Taylor Berg-Kirkpatrick, and Graham
  Neubig. 2022.
\newblock \href {https://openreview.net/forum?id=0RDcd5Axok} {Towards a unified
  view of parameter-efficient transfer learning}.
\newblock In \emph{International Conference on Learning Representations}.

\bibitem[{He et~al.(2021)He, Liu, Ye, Tan, Ding, Cheng, Low, Bing, and
  Si}]{he-etal-2021-effectiveness}
Ruidan He, Linlin Liu, Hai Ye, Qingyu Tan, Bosheng Ding, Liying Cheng, Jiawei
  Low, Lidong Bing, and Luo Si. 2021.
\newblock \href {https://doi.org/10.18653/v1/2021.acl-long.172} {On the
  effectiveness of adapter-based tuning for pretrained language model
  adaptation}.
\newblock In \emph{Proceedings of the 59th Annual Meeting of the Association
  for Computational Linguistics and the 11th International Joint Conference on
  Natural Language Processing (Volume 1: Long Papers)}, pages 2208--2222,
  Online. Association for Computational Linguistics.

\bibitem[{Houlsby et~al.(2019)Houlsby, Giurgiu, Jastrzebski, Morrone,
  De~Laroussilhe, Gesmundo, Attariyan, and Gelly}]{houlsby-etal-2019-parameter}
Neil Houlsby, Andrei Giurgiu, Stanislaw Jastrzebski, Bruna Morrone, Quentin
  De~Laroussilhe, Andrea Gesmundo, Mona Attariyan, and Sylvain Gelly. 2019.
\newblock Parameter-efficient transfer learning for {NLP}.
\newblock In \emph{International Conference on Machine Learning}, pages
  2790--2799. PMLR.

\bibitem[{Hu et~al.(2022)Hu, Shen, Wallis, Allen-Zhu, Li, Wang, Wang, and
  Chen}]{hu-etal-2022-lora}
Edward~J Hu, Yelong Shen, Phillip Wallis, Zeyuan Allen-Zhu, Yuanzhi Li, Shean
  Wang, Lu~Wang, and Weizhu Chen. 2022.
\newblock \href {https://openreview.net/forum?id=nZeVKeeFYf9} {Lo{RA}: Low-rank
  adaptation of large language models}.
\newblock In \emph{International Conference on Learning Representations}.

\bibitem[{Juki{\'c} and \v{S}najder(2023)}]{jukic-snajder-2023-smooth}
Josip Juki{\'c} and Jan \v{S}najder. 2023.
\newblock \href {https://aclanthology.org/2023.clasp-1.2} {Smooth sailing:
  Improving active learning for pre-trained language models with representation
  smoothness analysis}.
\newblock In \emph{Proceedings of the 2023 CLASP Conference on Learning with
  Small Data (LSD)}, pages 11--24, Gothenburg, Sweden. Association for
  Computational Linguistics.

\bibitem[{Karamcheti et~al.(2021)Karamcheti, Krishna, Fei-Fei, and
  Manning}]{karamcheti-etal-2021-mind}
Siddharth Karamcheti, Ranjay Krishna, Li~Fei-Fei, and Christopher Manning.
  2021.
\newblock \href {https://doi.org/10.18653/v1/2021.acl-long.564} {Mind your
  outliers! investigating the negative impact of outliers on active learning
  for visual question answering}.
\newblock In \emph{Proceedings of the 59th Annual Meeting of the Association
  for Computational Linguistics and the 11th International Joint Conference on
  Natural Language Processing (Volume 1: Long Papers)}, pages 7265--7281,
  Online. Association for Computational Linguistics.

\bibitem[{Karimi~Mahabadi et~al.(2021)Karimi~Mahabadi, Ruder, Dehghani, and
  Henderson}]{karimi-mahabadi-etal-2021-parameter}
Rabeeh Karimi~Mahabadi, Sebastian Ruder, Mostafa Dehghani, and James Henderson.
  2021.
\newblock \href {https://doi.org/10.18653/v1/2021.acl-long.47}
  {Parameter-efficient multi-task fine-tuning for transformers via shared
  hypernetworks}.
\newblock In \emph{Proceedings of the 59th Annual Meeting of the Association
  for Computational Linguistics and the 11th International Joint Conference on
  Natural Language Processing (Volume 1: Long Papers)}, pages 565--576, Online.
  Association for Computational Linguistics.

\bibitem[{Kim et~al.(2021)Kim, Shum, Susanj, and
  Hilgart}]{kim-etal-2021-revisiting}
Seungwon Kim, Alex Shum, Nathan Susanj, and Jonathan Hilgart. 2021.
\newblock \href {https://doi.org/10.18653/v1/2021.repl4nlp-1.11} {Revisiting
  pretraining with adapters}.
\newblock In \emph{Proceedings of the 6th Workshop on Representation Learning
  for NLP (RepL4NLP-2021)}, pages 90--99, Online. Association for Computational
  Linguistics.

\bibitem[{Kornblith et~al.(2019)Kornblith, Norouzi, Lee, and
  Hinton}]{kornblith-etal-2019-similarity}
Simon Kornblith, Mohammad Norouzi, Honglak Lee, and Geoffrey Hinton. 2019.
\newblock Similarity of neural network representations revisited.
\newblock In \emph{International Conference on Machine Learning}, pages
  3519--3529. PMLR.

\bibitem[{Lee et~al.(2022)Lee, Hwang, and Kim}]{lee-etal-2022-fad}
Jaeseong Lee, Seung-won Hwang, and Taesup Kim. 2022.
\newblock \href {https://aclanthology.org/2022.aacl-short.8} {{FAD}-{X}: Fusing
  adapters for cross-lingual transfer to low-resource languages}.
\newblock In \emph{Proceedings of the 2nd Conference of the Asia-Pacific
  Chapter of the Association for Computational Linguistics and the 12th
  International Joint Conference on Natural Language Processing (Volume 2:
  Short Papers)}, pages 57--64, Online only. Association for Computational
  Linguistics.

\bibitem[{Lewis and Gale(1994)}]{lewis-gale-1994-sequential}
David~D Lewis and William~A Gale. 1994.
\newblock A sequential algorithm for training text classifiers.
\newblock In \emph{SIGIR’94}, pages 3--12. Springer.

\bibitem[{Li and Liang(2021)}]{li-liang-2021-prefix}
Xiang~Lisa Li and Percy Liang. 2021.
\newblock \href {https://doi.org/10.18653/v1/2021.acl-long.353} {Prefix-tuning:
  Optimizing continuous prompts for generation}.
\newblock In \emph{Proceedings of the 59th Annual Meeting of the Association
  for Computational Linguistics and the 11th International Joint Conference on
  Natural Language Processing (Volume 1: Long Papers)}, pages 4582--4597,
  Online. Association for Computational Linguistics.

\bibitem[{Li and Roth(2002)}]{li-roth-2002-learning}
Xin Li and Dan Roth. 2002.
\newblock \href {https://aclanthology.org/C02-1150} {Learning question
  classifiers}.
\newblock In \emph{{COLING} 2002: The 19th International Conference on
  Computational Linguistics}.

\bibitem[{Mao et~al.(2022)Mao, Mathias, Hou, Almahairi, Ma, Han, Yih, and
  Khabsa}]{mao-etal-2022-unipelt}
Yuning Mao, Lambert Mathias, Rui Hou, Amjad Almahairi, Hao Ma, Jiawei Han,
  Scott Yih, and Madian Khabsa. 2022.
\newblock \href {https://doi.org/10.18653/v1/2022.acl-long.433} {{U}ni{PELT}: A
  unified framework for parameter-efficient language model tuning}.
\newblock In \emph{Proceedings of the 60th Annual Meeting of the Association
  for Computational Linguistics (Volume 1: Long Papers)}, pages 6253--6264,
  Dublin, Ireland. Association for Computational Linguistics.

\bibitem[{Margatina et~al.(2022)Margatina, Barrault, and
  Aletras}]{margatina-etal-2022-importance}
Katerina Margatina, Loic Barrault, and Nikolaos Aletras. 2022.
\newblock \href {https://doi.org/10.18653/v1/2022.acl-short.93} {On the
  importance of effectively adapting pretrained language models for active
  learning}.
\newblock In \emph{Proceedings of the 60th Annual Meeting of the Association
  for Computational Linguistics (Volume 2: Short Papers)}, pages 825--836,
  Dublin, Ireland. Association for Computational Linguistics.

\bibitem[{Margatina et~al.(2021)Margatina, Vernikos, Barrault, and
  Aletras}]{margatina-etal-2021-active}
Katerina Margatina, Giorgos Vernikos, Lo{\"\i}c Barrault, and Nikolaos Aletras.
  2021.
\newblock \href {https://doi.org/10.18653/v1/2021.emnlp-main.51} {Active
  learning by acquiring contrastive examples}.
\newblock In \emph{Proceedings of the 2021 Conference on Empirical Methods in
  Natural Language Processing}, pages 650--663, Online and Punta Cana,
  Dominican Republic. Association for Computational Linguistics.

\bibitem[{Mosbach et~al.(2021)Mosbach, Andriushchenko, and
  Klakow}]{mosbach-etal-2021-stability}
Marius Mosbach, Maksym Andriushchenko, and Dietrich Klakow. 2021.
\newblock \href {https://openreview.net/forum?id=nzpLWnVAyah} {On the stability
  of fine-tuning {BERT}: Misconceptions, explanations, and strong baselines}.
\newblock In \emph{International Conference on Learning Representations}.

\bibitem[{Pang and Lee(2004)}]{pang-lee-2004-sentimental}
Bo~Pang and Lillian Lee. 2004.
\newblock \href {https://doi.org/10.3115/1218955.1218990} {A sentimental
  education: Sentiment analysis using subjectivity summarization based on
  minimum cuts}.
\newblock In \emph{Proceedings of the 42nd Annual Meeting of the Association
  for Computational Linguistics ({ACL}-04)}, pages 271--278, Barcelona, Spain.

\bibitem[{Parovi{\'c} et~al.(2022)Parovi{\'c}, Glava{\v{s}}, Vuli{\'c}, and
  Korhonen}]{parovic-etal-2022-bad}
Marinela Parovi{\'c}, Goran Glava{\v{s}}, Ivan Vuli{\'c}, and Anna Korhonen.
  2022.
\newblock \href {https://doi.org/10.18653/v1/2022.naacl-main.130} {{BAD}-{X}:
  Bilingual adapters improve zero-shot cross-lingual transfer}.
\newblock In \emph{Proceedings of the 2022 Conference of the North American
  Chapter of the Association for Computational Linguistics: Human Language
  Technologies}, pages 1791--1799, Seattle, United States. Association for
  Computational Linguistics.

\bibitem[{Pfeiffer et~al.(2020)Pfeiffer, R{\"u}ckl{\'e}, Poth, Kamath,
  Vuli{\'c}, Ruder, Cho, and Gurevych}]{pfeiffer-etal-2020-adapterhub}
Jonas Pfeiffer, Andreas R{\"u}ckl{\'e}, Clifton Poth, Aishwarya Kamath, Ivan
  Vuli{\'c}, Sebastian Ruder, Kyunghyun Cho, and Iryna Gurevych. 2020.
\newblock \href {https://doi.org/10.18653/v1/2020.emnlp-demos.7}
  {{A}dapter{H}ub: A framework for adapting transformers}.
\newblock In \emph{Proceedings of the 2020 Conference on Empirical Methods in
  Natural Language Processing: System Demonstrations}, pages 46--54, Online.
  Association for Computational Linguistics.

\bibitem[{Pfeiffer et~al.(2023)Pfeiffer, Ruder, Vuli{\'c}, and
  Ponti}]{pfeiffer-etal-2023-modular}
Jonas Pfeiffer, Sebastian Ruder, Ivan Vuli{\'c}, and Edoardo~Maria Ponti. 2023.
\newblock Modular deep learning.
\newblock \emph{arXiv preprint arXiv:2302.11529}.

\bibitem[{Schr{\"o}der et~al.(2022)Schr{\"o}der, Niekler, and
  Potthast}]{schroder-etal-2022-revisiting}
Christopher Schr{\"o}der, Andreas Niekler, and Martin Potthast. 2022.
\newblock \href {https://doi.org/10.18653/v1/2022.findings-acl.172} {Revisiting
  uncertainty-based query strategies for active learning with transformers}.
\newblock In \emph{Findings of the Association for Computational Linguistics:
  ACL 2022}, pages 2194--2203, Dublin, Ireland. Association for Computational
  Linguistics.

\bibitem[{Sener and Savarese(2018)}]{sener-savarese-2018-active}
Ozan Sener and Silvio Savarese. 2018.
\newblock \href {https://openreview.net/forum?id=H1aIuk-RW} {Active learning
  for convolutional neural networks: A core-set approach}.
\newblock In \emph{International Conference on Learning Representations}.

\bibitem[{Settles(2009)}]{settles-2009-active}
Burr Settles. 2009.
\newblock \href
  {http://axon.cs.byu.edu/~martinez/classes/778/Papers/settles.activelearning.pdf}
  {Active learning literature survey}.
\newblock Computer sciences technical report, University of Wisconsin-Madison.

\bibitem[{Shelmanov et~al.(2021)Shelmanov, Puzyrev, Kupriyanova, Belyakov,
  Larionov, Khromov, Kozlova, Artemova, Dylov, and
  Panchenko}]{shelmanov-etal-2021-active}
Artem Shelmanov, Dmitri Puzyrev, Lyubov Kupriyanova, Denis Belyakov, Daniil
  Larionov, Nikita Khromov, Olga Kozlova, Ekaterina Artemova, Dmitry~V. Dylov,
  and Alexander Panchenko. 2021.
\newblock \href {https://doi.org/10.18653/v1/2021.eacl-main.145} {Active
  learning for sequence tagging with deep pre-trained models and {B}ayesian
  uncertainty estimates}.
\newblock In \emph{Proceedings of the 16th Conference of the European Chapter
  of the Association for Computational Linguistics: Main Volume}, pages
  1698--1712, Online. Association for Computational Linguistics.

\bibitem[{Socher et~al.(2013)Socher, Bauer, Manning, and
  Ng}]{socher-etal-2013-parsing}
Richard Socher, John Bauer, Christopher~D. Manning, and Andrew~Y. Ng. 2013.
\newblock \href {https://aclanthology.org/P13-1045} {Parsing with compositional
  vector grammars}.
\newblock In \emph{Proceedings of the 51st Annual Meeting of the Association
  for Computational Linguistics (Volume 1: Long Papers)}, pages 455--465,
  Sofia, Bulgaria. Association for Computational Linguistics.

\bibitem[{Srivastava et~al.(2014)Srivastava, Hinton, Krizhevsky, Sutskever, and
  Salakhutdinov}]{srivastava-etal-dropout}
Nitish Srivastava, Geoffrey Hinton, Alex Krizhevsky, Ilya Sutskever, and Ruslan
  Salakhutdinov. 2014.
\newblock \href {http://jmlr.org/papers/v15/srivastava14a.html} {Dropout: A
  simple way to prevent neural networks from overfitting}.
\newblock \emph{Journal of Machine Learning Research}, 15(56):1929--1958.

\bibitem[{Stephenson et~al.(2021)Stephenson, Padhy, Ganesh, Hui, Tang, and
  Chung}]{stephenson-etal-2021-geometry}
Cory Stephenson, Suchismita Padhy, Abhinav Ganesh, Yue Hui, Hanlin Tang, and
  SueYeon Chung. 2021.
\newblock \href {https://openreview.net/forum?id=V8jrrnwGbuc} {On the geometry
  of generalization and memorization in deep neural networks}.
\newblock In \emph{International Conference on Learning Representations}.

\bibitem[{Toneva et~al.(2019)Toneva, Sordoni, des Combes, Trischler, Bengio,
  and Gordon}]{toneva-etal-2019-empirical}
Mariya Toneva, Alessandro Sordoni, Remi~Tachet des Combes, Adam Trischler,
  Yoshua Bengio, and Geoffrey~J. Gordon. 2019.
\newblock \href {https://openreview.net/forum?id=BJlxm30cKm} {An empirical
  study of example forgetting during deep neural network learning}.
\newblock In \emph{International Conference on Learning Representations}.

\bibitem[{Yu et~al.(2022)Yu, Kong, Zhang, Zhang, and
  Zhang}]{yu-etal-2022-actune}
Yue Yu, Lingkai Kong, Jieyu Zhang, Rongzhi Zhang, and Chao Zhang. 2022.
\newblock \href {https://doi.org/10.18653/v1/2022.naacl-main.102} {{A}c{T}une:
  Uncertainty-based active self-training for active fine-tuning of pretrained
  language models}.
\newblock In \emph{Proceedings of the 2022 Conference of the North American
  Chapter of the Association for Computational Linguistics: Human Language
  Technologies}, pages 1422--1436, Seattle, United States. Association for
  Computational Linguistics.

\bibitem[{Yuan et~al.(2020)Yuan, Lin, and Boyd-Graber}]{yuan-etal-2020-cold}
Michelle Yuan, Hsuan-Tien Lin, and Jordan Boyd-Graber. 2020.
\newblock \href {https://doi.org/10.18653/v1/2020.emnlp-main.637} {Cold-start
  active learning through self-supervised language modeling}.
\newblock In \emph{Proceedings of the 2020 Conference on Empirical Methods in
  Natural Language Processing (EMNLP)}, pages 7935--7948, Online. Association
  for Computational Linguistics.

\bibitem[{Zhang et~al.(2021)Zhang, Wu, Katiyar, Weinberger, and
  Artzi}]{zhang-etal-2021-revisiting}
Tianyi Zhang, Felix Wu, Arzoo Katiyar, Kilian~Q Weinberger, and Yoav Artzi.
  2021.
\newblock \href {https://openreview.net/forum?id=cO1IH43yUF} {Revisiting
  few-sample {BERT} fine-tuning}.
\newblock In \emph{International Conference on Learning Representations}.

\bibitem[{Zhang et~al.(2015)Zhang, Zhao, and LeCun}]{zhang-etal-2015-character}
Xiang Zhang, Junbo Zhao, and Yann LeCun. 2015.
\newblock Character-level convolutional networks for text classification.
\newblock \emph{Advances in neural information processing systems}, 28.

\end{thebibliography}
\bibliographystyle{acl_natbib}

\clearpage
\appendix

\section{Reproducibility}

\subsection{Dataset statistics}

The sizes of the datasets per split are provided in \Cref{tab:dataset-stats}. Predominantly, the datasets encompass texts in English.

\begin{table}[t!]
\small
\centering
\begin{tabular}{lrrrr}
\toprule
& \textsc{train} & \textsc{val} & \textsc{test} & \textsc{total} \\
\midrule
\subj & $7,000$ & $1,000$ & $2,000$ & $10,000$ \\
\sst & $6,647$ & $868$ & $1,425$ & $8,940$ \\
\trec & $4,881$ & $452$ & $500$ & $5,833$ \\
\agn & $20,000$ & $7,600$ & $7,600$ & $35,200$ \\
\bottomrule
\end{tabular}
\caption{Dataset sizes by splits. Although we do not use a validation set (\textsc{val}) in our experiments, we report its size for completeness. For the \agn{} dataset, we performed uniform subsampling to ensure the computational feasibility of the experiments.} 
\label{tab:dataset-stats}
\end{table}

\subsection{Adapters}
\label{app:ada}
We use the implementation of adapters from AdapterHub \cite{pfeiffer-etal-2020-adapterhub}.
\begin{description}
\item[\adapter{}] We set reduction factor to $16$ and use \textit{swish} function as nonlinearity.
\item[\lora] We include \lora{} to the self-attention weights, intermediate, and output MLP weights of a model. We set the rank of the \lora{} layer and the scaling factor $\alpha$ to $8$.
\item[\pt] We use \textit{tanh} activation for \pt{}, with prefix length set to $30$ and bottleneck size of $512$.
\item[\uni] We use \adapter{}, \lora{}, and \pt{} as components of \uni{} with the same hyperparameters as described for individual components. The only exception is that we set the prefix length for \pt{} to $10$ instead of $30$.
\end{description}

\subsection{AL methods}
\begin{description}
\item[\textsc{mc}] In experiments, we use ten inference cycles to approximate the entropy of the output via Monte-Carlo dropout sampling.
\item[\textsc{cs}] We use the [\textsc{cls}] token representation from the Transformer's penultimate layer. We follow the greedy method described in the original work \cite{sener-savarese-2018-active}
\end{description}

\subsection{Preprocessing}

We undertake a few pre-processing steps: convert all tokens to lowercase, eliminate non-alphanumeric tokens, and limit the token sequence to a maximum length of $200$.

\subsection{Hyperparameters}
\label{app:hyper}

We use a fixed learning rate of $2 \times 10^{-5}$ for FFT and $10^{-4}$ for adapters. Additionally, we set the gradient clipping to $1$ during training. In our implementation of TAPT, we randomly mask $15\%$ of tokens for both FFT models and adapters and train the model for 50 epochs with the learning rate set to $10^{-5}$.

\subsection{Computing infrastructure}
We conducted our experiments on $4 \times$ \textit{AMD Ryzen Threadripper 3970X 32-Core Processors} and $4 \times$ \textit{NVIDIA GeForce RTX 3090} GPUs with $24$GB of RAM. We used \textit{PyTorch} version $1.9.0$ and CUDA $11.4$.

\subsection{Average runtime}

We report the average runtime of experiments in \Cref{tab:runtime}.

\begin{table}
\setlength{\tabcolsep}{4pt} 
\centering
\small
\begin{tabular}{lccccc}
\toprule
& FFT & \adapter{} & \lora{} & \pt{} & \uni{} \\
\midrule
\subj{} & $40.8$ & $4.3$ & $3.2$ & $4.1$ & $6.2$ \\
\trec{} & $68.4$ & $4.9$ & $5.4$ & $5.9$ & $8.4$ \\
\sst{} & $43.9$ & $5.1$ & $5.0$ & $4.9$ & $7.6$ \\
\agn{} & $72.1$ & $7.3$ & $6.1$ & $4.4$ & $9.3$ \\
\bottomrule
\end{tabular}
\caption{Experiment duration in minutes for all models across datasets. We report the average runtime over five different runs and five different sampling methods (five AL methods and random sampling).}
\label{tab:runtime}
\end{table}

\section{Additional Results}
\label{app:exp}

\begin{figure*}[h]
    \centering
    \begin{subfigure}[b]{.195\linewidth}
        \includegraphics[width=\linewidth]{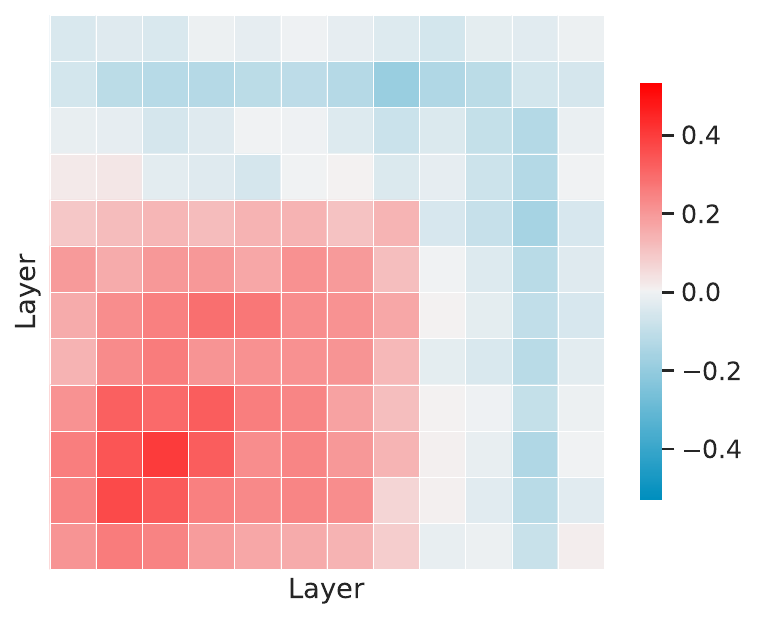}
        \caption{\trec{}; \ent{}}
    \end{subfigure}
    \begin{subfigure}[b]{.195\linewidth}
        \includegraphics[width=\linewidth]{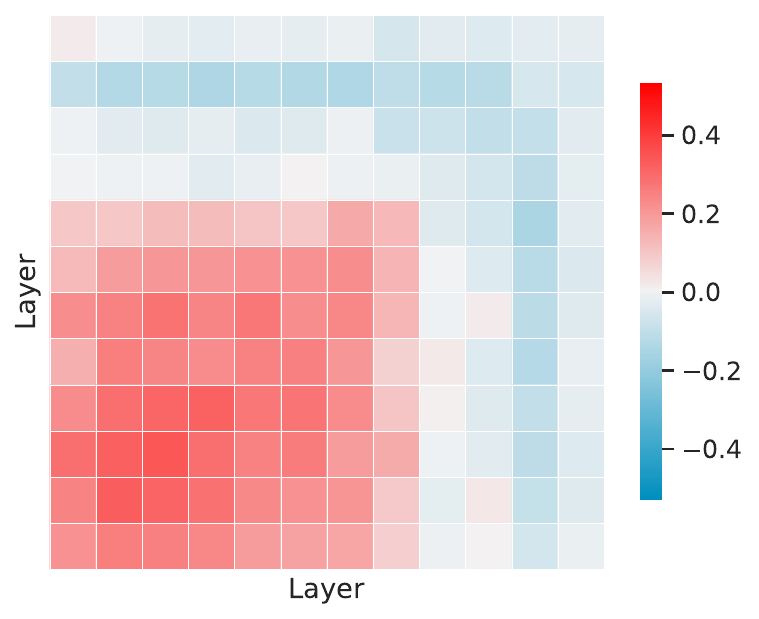}
        \caption{\trec{}; \mc{}}
    \end{subfigure}
    \begin{subfigure}[b]{0.195\linewidth}
        \includegraphics[width=\linewidth]{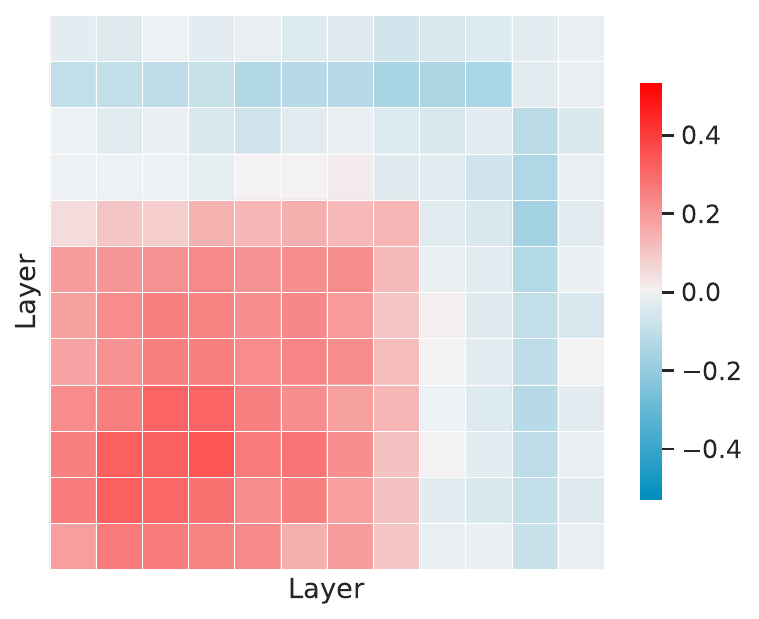}
        \caption{\trec{}; \cs{}}
    \end{subfigure}
    \begin{subfigure}[b]{0.195\linewidth}
        \includegraphics[width=\linewidth]{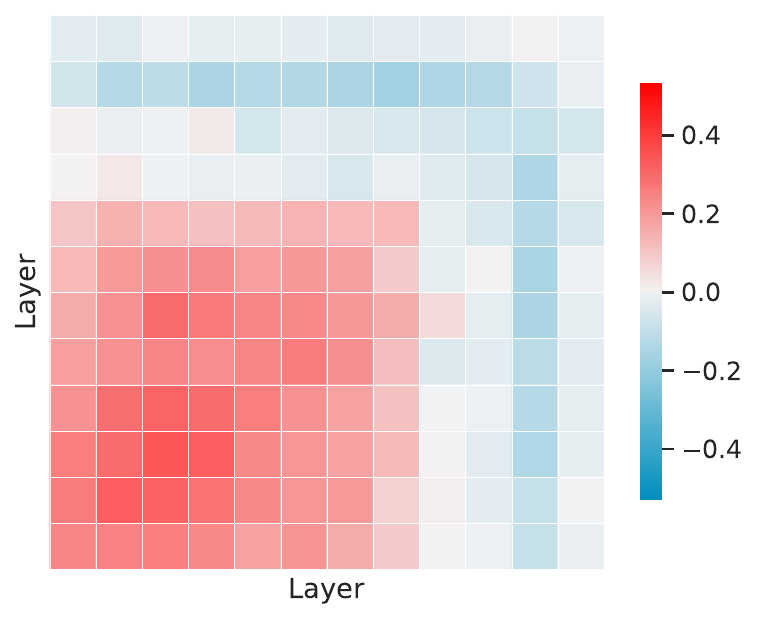}
        \caption{\trec{}; \dal{}}
    \end{subfigure}
    \begin{subfigure}[b]{0.195\linewidth}
        \includegraphics[width=\linewidth]{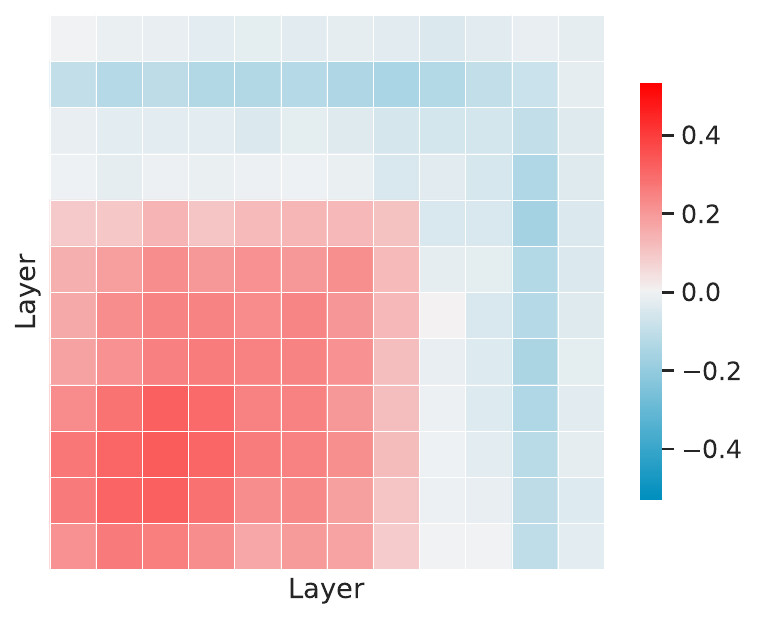}
        \caption{\trec{}; \rnd{}}
    \end{subfigure}

    \begin{subfigure}[b]{.195\linewidth}
        \includegraphics[width=\linewidth]{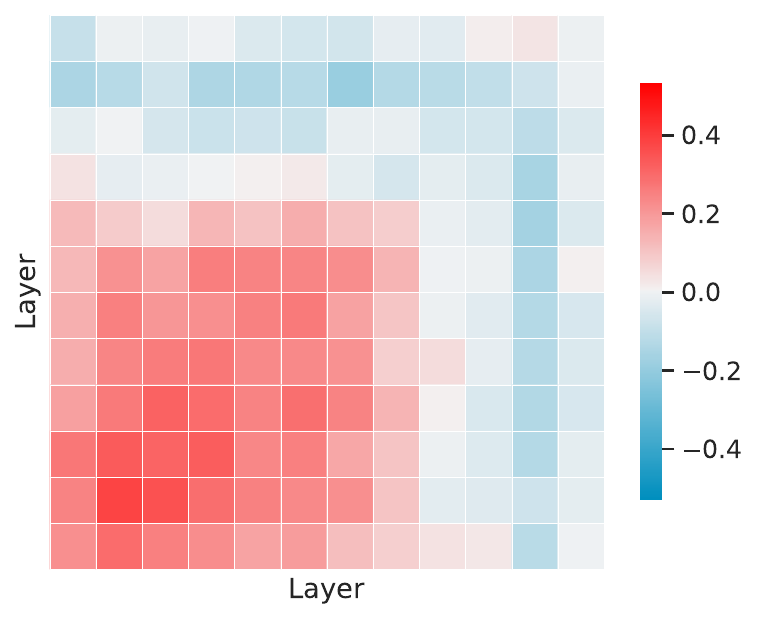}
        \caption{\sst{}; \ent{}}
    \end{subfigure}
    \begin{subfigure}[b]{.195\linewidth}
        \includegraphics[width=\linewidth]{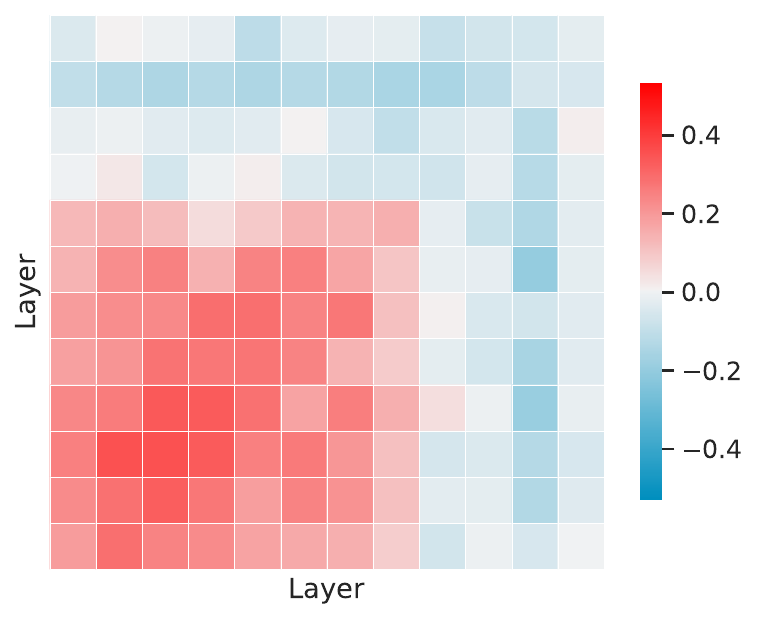}
        \caption{\sst{}; \mc{}}
    \end{subfigure}
    \begin{subfigure}[b]{0.195\linewidth}
        \includegraphics[width=\linewidth]{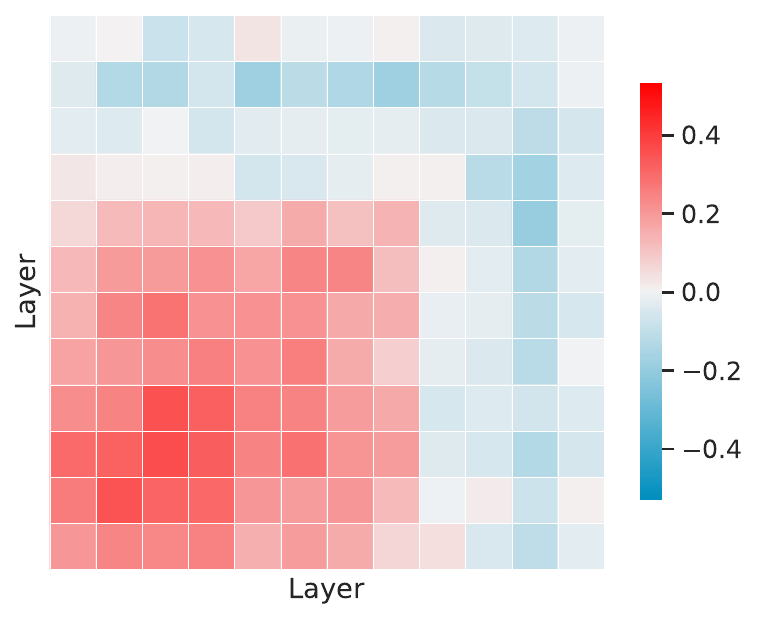}
        \caption{\sst{}; \cs{}}
    \end{subfigure}
    \begin{subfigure}[b]{0.195\linewidth}
        \includegraphics[width=\linewidth]{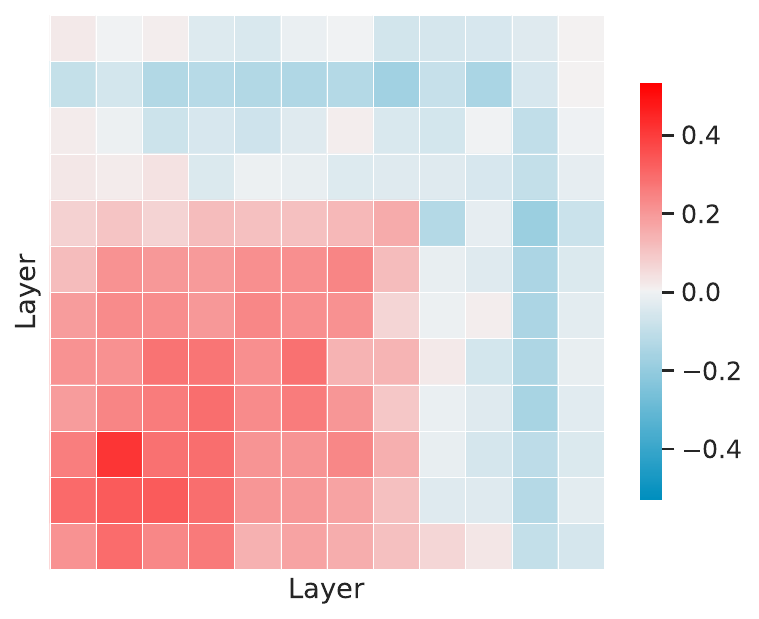}
        \caption{\sst{}; \dal{}}
    \end{subfigure}
    \begin{subfigure}[b]{0.195\linewidth}
        \includegraphics[width=\linewidth]{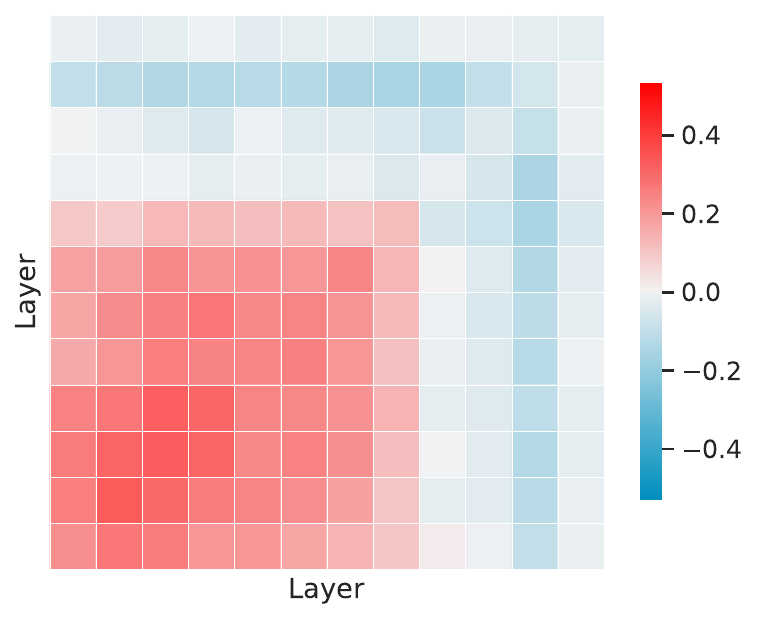}
        \caption{\sst{}; \rnd{}}
    \end{subfigure}

    \begin{subfigure}[b]{.195\linewidth}
        \includegraphics[width=\linewidth]{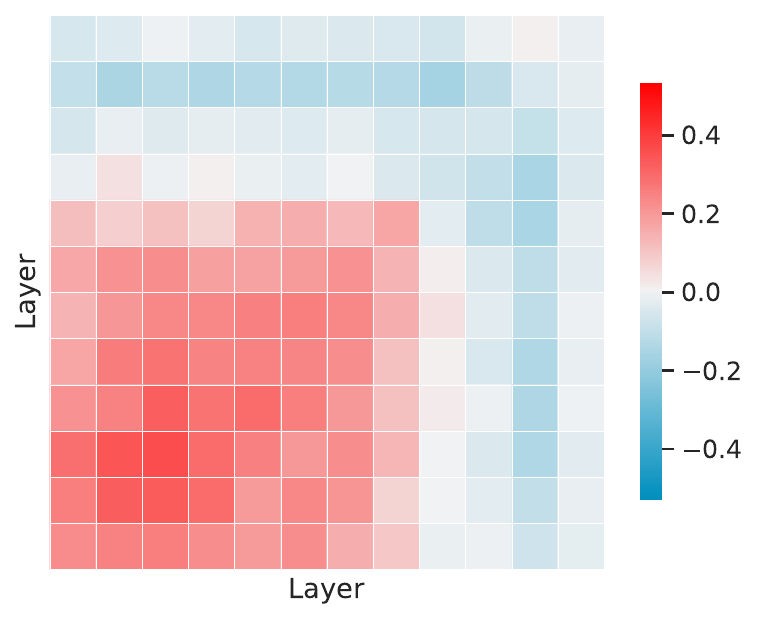}
        \caption{\agn{}; \ent{}}
    \end{subfigure}
    \begin{subfigure}[b]{.195\linewidth}
        \includegraphics[width=\linewidth]{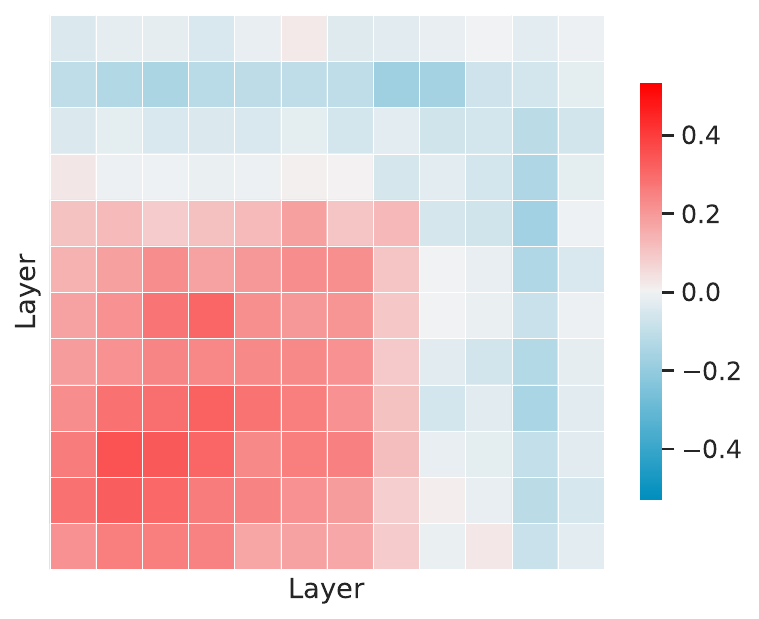}
        \caption{\agn{}; \mc{}}
    \end{subfigure}
    \begin{subfigure}[b]{0.195\linewidth}
        \includegraphics[width=\linewidth]{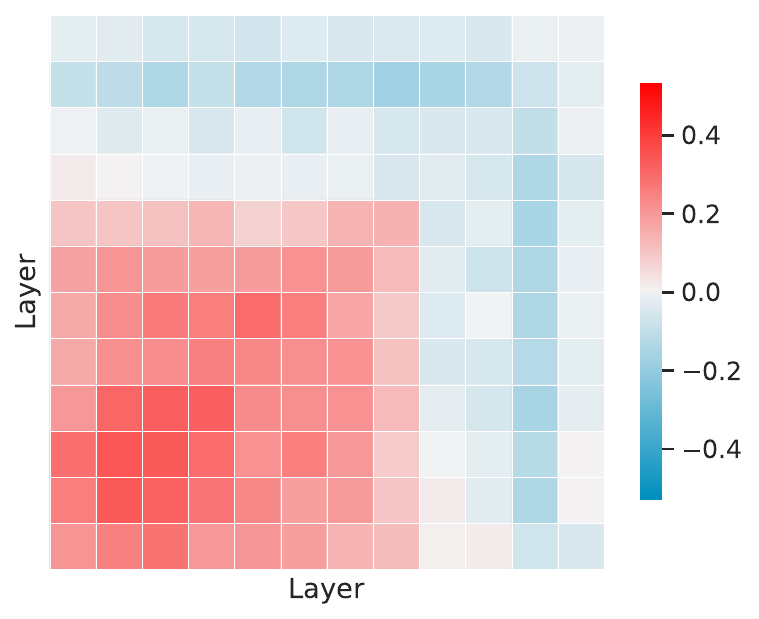}
        \caption{\agn{}; \cs{}}
    \end{subfigure}
    \begin{subfigure}[b]{0.195\linewidth}
        \includegraphics[width=\linewidth]{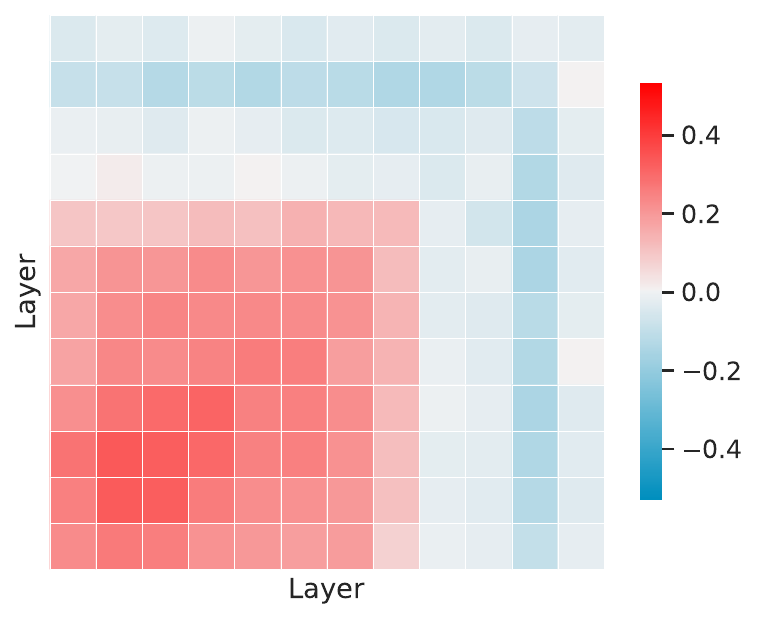}
        \caption{\agn{}; \dal{}}
    \end{subfigure}
    \begin{subfigure}[b]{0.195\linewidth}
        \includegraphics[width=\linewidth]{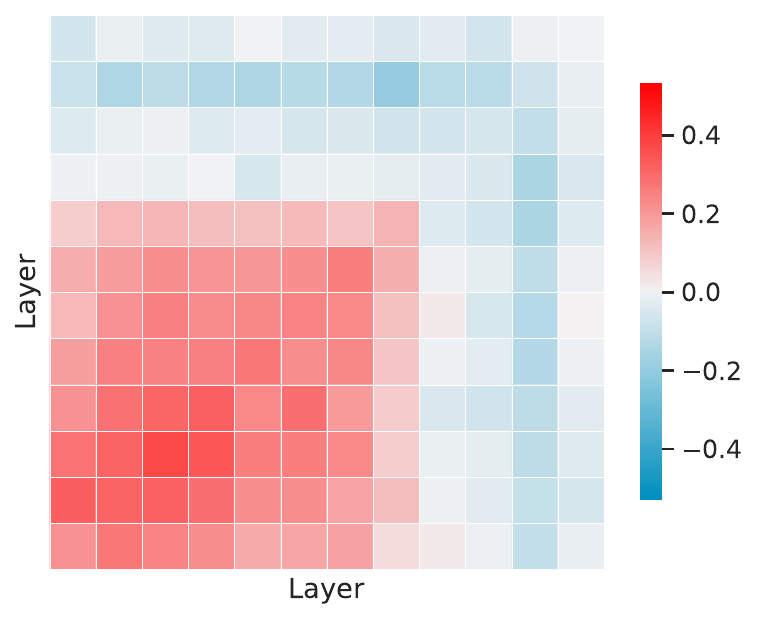}
        \caption{\agn{}; \rnd{}}
    \end{subfigure}

\caption{Layerwise difference in representation similarity for the \uni{} adapter and the FFT model on \trec{}, \sst{}, and \agn{}. The differences are computed as $\mathrm{CKA}(\textit{adapter}, \textit{base}) - \mathrm{CKA}(\textit{FFT}, \textit{base})$, where \textit{base} is the corresponding pre-trained \bert{} model. Warm colors (positive values) illustrate layer pairs that demonstrate higher similarity to the base model with the adapter than with FFT. Conversely, cool colors (negative values) represent layer pairs that are more similar to the base model when using the FFT model. Best viewed on a computer screen.}
\label{fig:repr_diff_app}
\end{figure*}

We report the results that were omitted from the main part of the paper due to space constraints. \Cref{tab:aucs-app} shows AUC scores for different combinations of AL methods and adapters, complementing the relative improvement scores as AUC represents absolute scores for each configuration. In \Cref{fig:repr_diff_app}, we display the difference in similarities of adapters and FFT compared to their base models on the remaining three datasets.

\begin{table*}[htb!]
\centering
\small
\begin{tabular}{lr|rrrrr|rrrrr}
\toprule
& & 
\multicolumn{5}{c}{without TAPT} & \multicolumn{5}{c}{with TAPT} \\
\cmidrule(lr){3-7} \cmidrule(lr){8-12} 
 & & \rnd{} & \ent{} & \mc{} & \cs{} & \dal{} & \rnd{} & \ent{} & \mc{} & \cs{} & \dal{} \\
\midrule
\multirow{5}{*}{\rotatebox[origin=c]{90}{\subj}}
& FFT & $.928$ & $.931$ & $.932$ & $.932$ & $.934$ & $.938$ & $.947$ & $.947$ & $.947$ & $.946$\\
& \adapter{} & $.926$ & $.934$ & $.933$ & $.933$ & $.932$ & $.934$ & $.943$ & $.944$ & $.941$ & $.938$
\\
& \lora{} & $.929$ & $.938$ & $.937$ & $.935$ & $.934$ & $.935$ & $.946$ & $.945$ & $.943$ & $.942$\\
& \pt & $.936$ & $.942$ & $.943$ & $.943$ & $.943$ & $.940$ & $.951$ & $.951$ & $.950$ & $.949$\\
& \uni{} & $.934$ & $.943$ & $\textbf{.944}$ & $.943$ & $.942$ & $.943$ & $.952$ & $\textbf{.953}$ & $.952$ & $.952$\\
\midrule
\multirow{5}{*}{\rotatebox[origin=c]{90}{\trec}}
& FFT & $.810$ & $.812$ & $.814$ & $.817$ & $.816$ & $.818$ & $.847$ & $.851$ & $.844$ & $.847$\\
& \adapter{} & $.804$ & $.809$ & $.818$ & $.831$ & $.820$ & $.820$ & $.842$ & $.846$ & $.834$ & $.848$\\
& \lora{} & $.750$ & $.775$ & $.766$ & $.762$ & $.752$ & $.764$ & $.824$ & $.820$ & $.821$ & $.781$\\
& \pt{} & $.847$ & $.861$ & $.863$ & $.857$ & $.861$ & $.862$ & $.896$ & $.893$ & $.890$ & $.895$\\
& \uni{} & $.877$ & $.894$ & $.897$ & $.887$ & $\textbf{.902}$ & $.896$ & $.927$ & $\textbf{.931}$ & $.925$ & $.921$\\
\midrule
\multirow{5}{*}{\rotatebox[origin=c]{90}{\sst}}
& FFT & $.787$ & $.787$ & $.789$ & $.779$ & $.788$ & $.792$ & $.809$ & $.808$ & $.808$ & $.807$\\
& \adapter{} & $.800$ & $.803$ & $.810$ & $.805$ & $.801$ & $.812$ & $.819$ & $.818$ & $.817$ & $.814$\\
& \lora{} & $.798$ & $.798$ & $.799$ & $.811$ & $.804$ & $.806$ & $.813$ & $.810$ & $.812$ & $.809$\\
& \pt{} & $.847$ & $.854$ & $.856$ & $\textbf{.864}$ & $.852$ & $.868$ & $\textbf{.888}$ & $.887$ & $.886$ & $.885$\\
& \uni{} & $.836$ & $.842$ & $.843$ & $.843$ & $.837$ & $.871$ & $.882$ & $.884$ & $.882$ & $.881$
\\
\midrule
\multirow{5}{*}{\rotatebox[origin=c]{90}{\agn}}
& FFT & $.860$ & $.862$ & $.864$ & $.861$ & $.873$ & $.869$ & $.887$ & $.872$ & $.881$ & $.871$\\
& \adapter{} & $.871$ & $.881$ & $.877$ & $.873$ & $.879$ & $.882$ & $.896$ & $.893$ & $.891$ & $.891$\\
& \lora{}& $.860$ & $.863$ & $.863$ & $.869$ & $.862$ & $.868$ & $.872$ & $.881$ & $.877$ & $.871$\\
& \pt{} & $.875$ & $.882$ & $.878$ & $.880$ & $.879$ & $.886$ & $.890$ & $.902$ & $.897$ & $.896$\\
& \uni{} & $.875$ & $.884$ & $\textbf{.887}$ & $.886$ & $\textbf{.887}$
 & $.887$ & $\textbf{.908}$ & $.904$ & $.900$ & $.896$\\
\bottomrule
\end{tabular}
\caption{AUC scores for AL methods with different adapters shown separately without TAPT and with TAPT. We include random sampling for comparison with AL methods. Values in \textbf{bold} denote the best result for a particular dataset within the same regime (with or without TAPT).}
\label{tab:aucs-app}
\end{table*}

\end{document}